\documentclass{article}

\usepackage{arxiv}

\usepackage[utf8]{inputenc} 
\usepackage[T1]{fontenc}    
\usepackage{hyperref}       
\usepackage{url}            
\usepackage{booktabs}       
\usepackage{amsfonts}       
\usepackage{nicefrac}       
\usepackage{microtype}      
\usepackage{lipsum}		
\usepackage{graphicx}
\usepackage{natbib}
\usepackage{doi}

\usepackage{multirow}
\usepackage{booktabs}
\usepackage{amsmath}
\usepackage{amssymb}

\usepackage{geometry}
\usepackage{times}
\usepackage{array}
\usepackage{varwidth}
\usepackage{latexsym}
\usepackage{url}
\usepackage{lscape}
\usepackage{color,soul}
\usepackage{graphicx}
\usepackage{caption}
\usepackage{subcaption}
\usepackage{enumitem}
\usepackage[framed,numbered]{matlab-prettifier}
\usepackage{threeparttable}
\usepackage[T1]{fontenc}
\DeclareUnicodeCharacter{00A0}{ }

\title{Large Language Models, scientific knowledge and factuality: \\ A framework to streamline human expert evaluation}

\author{
  \textbf{Magdalena Wysocka\textsuperscript{1}\thanks{Corresponding author: magdalena.wysocka@cruk.manchester.ac.uk}},
  \textbf{Oskar Wysocki\textsuperscript{1,2}},
  \textbf{Maxime Delmas\textsuperscript{2}},
\\
  \textbf{Vincent Mutel\textsuperscript{3}},
  \textbf{Andr\'e Freitas\textsuperscript{1,2,4}}
\\
\\
  \textsuperscript{1}National Biomarker Centre, CRUK-MI, Univ. of Manchester, United Kingdom\\
  \textsuperscript{2}Idiap Research Institute, Switzerland\\
  \textsuperscript{3}Inflamalps SA, Monthey, Switzerland\\
  \textsuperscript{4}Department of Computer Science, Univ. of Manchester, United Kingdom
\\
  \small{
  }
}

\renewcommand{\shorttitle}{A preprint}

\hypersetup{
pdftitle={A template for the arxiv style},
pdfsubject={q-bio.NC, q-bio.QM},
pdfauthor={David S.~Hippocampus, Elias D.~Striatum},
pdfkeywords={First keyword, Second keyword, More},
}

\begin{document}
\maketitle

\begin{abstract}
The paper introduces a framework for the evaluation of the encoding of factual scientific knowledge, designed to streamline the manual evaluation process typically conducted by domain experts.  
Inferring over and extracting information from Large Language Models (LLMs) trained on a large corpus of scientific literature can potentially define a step change in biomedical discovery, reducing the barriers for accessing and integrating existing medical evidence. This work explores the potential of LLMs for dialoguing with biomedical background knowledge, using the context of antibiotic discovery. 
The framework involves of three evaluation steps, each assessing different aspects sequentially: fluency, prompt alignment, semantic coherence, factual knowledge, and specificity of the generated responses. By splitting these tasks between non-experts and experts, the framework reduces the effort required from the latter.
The work provides a systematic assessment on the ability of eleven state-of-the-art models LLMs, including ChatGPT, GPT-4 and Llama 2, in two prompting-based tasks: chemical compound definition generation and chemical compound-fungus relation determination. 
Although recent models have improved in fluency, factual accuracy is still low and models are biased towards over-represented entities. The ability of LLMs to serve as biomedical knowledge bases is questioned, and the need for additional systematic evaluation frameworks is highlighted. 
While LLMs are currently not fit for purpose to be used as biomedical factual knowledge bases in a zero-shot setting, there is a promising emerging property in the direction of factuality as the models become domain specialised, scale-up in size and level of human feedback.   
\end{abstract}

\section{Introduction}

Despite comprehensive linguistic interpretation properties, LLMs have inherent operational limitations, most notably hallucinations \cite{Bender2021, Ji2023hallucination}, i.e. divergence from factual knowledge and spurious inference, despite its fluency and syntactic correctness \cite{mahowald2023dissociating, weidingerTaxonomyRisksPosed2022}. 
This elicits the need to establish the ability of these models to encode and preserve factual knowledge as well as to define methodologies for the critical evaluation of its representation properties \cite{wysocki2022transformers, jullien2022transformers, rozanova2023interventional}.

In this work we introduce an evaluation framework, which is designed to streamline the manual evaluation process typically conducted by domain experts. Despite the advancement and implementation of automated evaluation methods \cite{li2024leveraging}, the necessity for manual, human-led verification persists \cite{Bavaresco2024}. Users engaged in the biomedical field are likely to continue performing assessments of models within the context of new tasks, using examples they are well-acquainted with to ensure the model's accuracy and reliability. The framework, therefore, is not meant to replace human evaluation but to guide users on key aspects to consider when conducting their assessments, acting as a roadmap for thorough and effective manual evaluation. Being transferable to other biomedical domains, it enables the recognition of, i.a., biases in LLM outputs towards prevalent topics, challenges in identifying rare entities, deviations from intended contexts, and significant performance variability influenced by prompt design.

Using the framework, we systematically assess the biological relational knowledge encoded in a broad spectrum of state-of-the-art LLMs, namely: GPT-2, GPT-3, GPT-neo, Bloom, BioGPT, BioGPT-Large, ChatGPT, GPT-4 and variants of Llama 2 with 7B, 13B, and 70B parameters. 
Recently, artificial intelligence (AI) algorithms have significantly facilitated the acceleration of antibiotic discovery through high-throughput drug screening, with a particular emphasis on the identification of novel antibiotic compounds \cite{jablonka2024leveraging, guo2023can, TORRES201930, lluka2023antibiotic, ruiz2022rational, david2021artificial, melo2021accelerating}. Motivated by this progress, we focus on systematically determining the ability of LLMs to capture fundamental entities (fungi, chemicals and antibiotic properties), their relations and supporting facts, which are fundamental for delivering inference in the context of downstream biomedical inference. For instance, we aim to answer the question on whether LLMs can faithfully capture biological domain-specific factual knowledge such as: \textit{``\textit{Aspergillus fumigatus} produces festuclavine"}; \textit{``Cycloclavine is biosynthesised by \textit{Aspergillus japonicus}"}.

This overarching question is translated into the following specific research questions (RQs):

RQ1: Do large language models encode biomedical domain knowledge at entity level (e.g. fungi, chemical, antibiotic properties) and at a relational level? 

RQ2: Are there significant differences between different models in their ability of encoding domain knowledge? 

RQ3: Which model performs better in faithfully encoding biological relations?

RQ4: How do we evaluate the limitations of LLMs in factual knowledge extraction?

This study is organised as follows: first, we define our contribution and link it with the existing frameworks and other investigations within the biomedical domain. Then, within the Materials and methods section, we present a framework for the qualitative evaluation of the biomedical knowledge embedded in LLMs. 
In the Results section, we apply the proposed framework to eleven LLMs and evaluate them for ten chemical compounds with a detailed description of the generated responses. We then discuss our findings in the broader context of reported LLMs' limitations. Finally, we conclude the paper with a summary of the contributions and with future perspectives on the role of LLMs in biomedical inference.

\section{Related Work}
LLMs achieve state-of-the-art performance in multiple Natural Language Inference and Understanding tasks, frequently outperforming previous baselines in few-shot settings \cite{brown2020language}. The knowledge extracted by prompting these models has been widely discussed in the context of its factuality \cite{survey_lee2023,Bavaresco2024}. Since the first demonstration by Petroni et al. \cite{petroniLanguageModelsKnowledge2019} on Masked Language Models, several studies have examined the mechanisms and biases underlying these outputs, questioning their applicability beyond their potential as knowledge bases.

Howard et al. \cite{howard_chatgpt_2023} examined the implications of generative AI for antimicrobial advisory for addressing infection questions and outlined the limitations of implementing ChatGPT for interventions due to deficits in situational awareness, inference, and consistency. Li et al. \cite{Li2023JianningChatGPT} reported that ChatGPT performs moderately or poorly in several biomedical tests and is unreliable in actual clinical use. Furthermore, integrating LLMs into clinical practice is associated with numerous obstacles, such as deficits in situational awareness, lack of reasoning control mechanisms, and consistency \cite{WangNLPmedicine2020}. While the use of natural language processing in healthcare is not novel, LLMs have elicited intense debate regarding its potential opportunities and challenges in healthcare \cite{nori2023capabilities}.

This paper builds upon findings from several key studies to address the limitations and potentials of LLMs in encoding scientific domain knowledge. Zhao et al. \cite{zhaoCalibrateUseImproving2021}, Kassner, Krojer, and Schutze \cite{kassnerArePretrainedLanguage2020} and Wysocki et al. \cite{wysocki2022transformers} emphasise the inconsistency of language models and their tendency to replicate prevalent responses, a form of co-occurrence bias also noted in LLMs by Kandpal et al. \cite{kandpalLargeLanguageModels2023} and Kang and Choi \cite{kangImpactCooccurrenceFactual2023}. This tendency undermines the models' ability to generalise, a concern supported by Razeghi et al. \cite{razeghiImpactPretrainingTerm2022}, who report an accuracy disparity tied to the frequency of terms in training data. Biderman et al. \cite{bidermanEmergentPredictableMemorization2023}, Power et al. \cite{powerGrokkingGeneralizationOverfitting2022} and Tirumala et al. \cite{tirumalaMemorizationOverfittingAnalyzing2022} explore delicate balance between memorisation's beneficial role in preventing the hallucinations, versus outputting entire sequences from their training data verbatim. This is a key safety concern, particularly in preventing the unintended disclosure of sensitive or proprietary information from the training data \cite{carlini2021extracting}. Wang et al. \cite{wangRetrieveWhatYou} and Delmas, Wysocka, and Freitas \cite{delmasRelationExtractionUnderexplored2023} remark on LLMs' underperformance when compared to supervised settings, positing retrieval-augmentation as a potential remedy for knowledge-dense tasks. Together, these studies highlight the need for strategies to refine LLMs' bias and factuality, enhancing their application in biomedical domain.

The essential requirement for the biomedical utility of LLMs is the factual accuracy of the generated text.
Standard N-Gram matching metrics, including BLEU \cite{papineni-etal-2002-bleu}, ROUGE \cite{lin-2004-rouge}, and token-level F1, have been found to poorly correlate with factual consistency \cite{maynez-etal-2020-faithfulness, honovich-etal-2021-q2}. Given the cumbersome nature of human evaluation for factuality, there is a rising inclination towards employing LLMs for automating the evaluation of model output. Several studies suggest that LLMs can effectively serve as evaluators, often aligning well with human assessments \cite{liu-etal-2023-g, zheng2023judgingllmasajudgemtbenchchatbot, chen-etal-2023-exploring-use, törnberg2023chatgpt4outperformsexpertscrowd, huang-etal-2024-chatgpt-rates, naismith-etal-2023-automated, Gilardi_2023, kocmi-federmann-2023-large, verga-2024}, though there are some exceptions \cite{wang-etal-2023-chatgpt, wu2023stylesubstanceevaluationbiases, hada-etal-2024-large, pavlovic-poesio-2024-effectiveness}. In contrast, other research highlights notable deficiencies in the performance of LLMs as evaluators \cite{koo2023benchmarkingcognitivebiaseslarge, zeng2024evaluating, baris-schlicht-etal-2024-pitfalls}, and some studies fail to benchmark LLMs against human judgments \cite{jiang-etal-2023-llm, landwehr-etal-2023-memories}.
For summary LLM-based evaluation methodologies, we refer to \cite{li2024leveraging}. Bavaresco et al. conclude that LLMs are not yet equipped to systematically replace human judges in NLP, reinforcing our argument for the need for human evaluation \cite{Bavaresco2024}.

Existing frameworks for evaluating the factuality of LLMs highlight various approaches and their effectiveness. Luo et al. proposes a systematic framework using diverse, high-coverage questions generated from Knowledge Graphs \cite{luo2023systematic}. Hendrycks et al. measures how well text models learn and apply knowledge encountered during pretraining, assessing language understanding across 57 subjects of varying difficulty \cite{hendrycks2020measuring}. The FRANK survey evaluates the faithfulness metrics for summarisation and compares the correlations of these metrics with human judgments, introducing a typology of errors for factual consistency \cite{pagnoni-etal-2021-understanding}. Rashkin et al. suggests that binary labeling is more beneficial for practical applications where filtering out unfaithful predictions is necessary, aligning with recommendations for human evaluation of attribution in text generation \cite{rashkin2021}. Sun et al. provides a comprehensive study on the trustworthiness of LLMs, establishing benchmarks and principles for different dimensions of trustworthiness \cite{sun2024trustllm}.

In the bioinformatics domain, Yin et al. provides a thorough analysis of the limitations of LLMs in complex bioinformatics tasks \cite{yin2024evaluation}. Piccolo et al. evaluates LLMs in bioinformatics programming, focusing on their ability to handle domain-specific coding tasks \cite{piccolo2023many}. Park et al. conducts a comparative evaluation of LLMs for automatically extracting knowledge from scientific literature to understand protein interactions and pathway knowledge \cite{park2023comparative}. Our work introduces a unique evaluation framework that not only rigorously tests LLMs' ability to encode and apply biomedical knowledge in antibiotic discovery but also significantly reduces the effort required from human experts. This focus on minimizing human evaluation sets our approach apart from previous studies, providing new insights and actionable recommendations to improve factual accuracy and domain-specific reliability.

\section{Materials and methods}

\begin{figure}[h!]
\centering
\includegraphics[width= .99\textwidth]{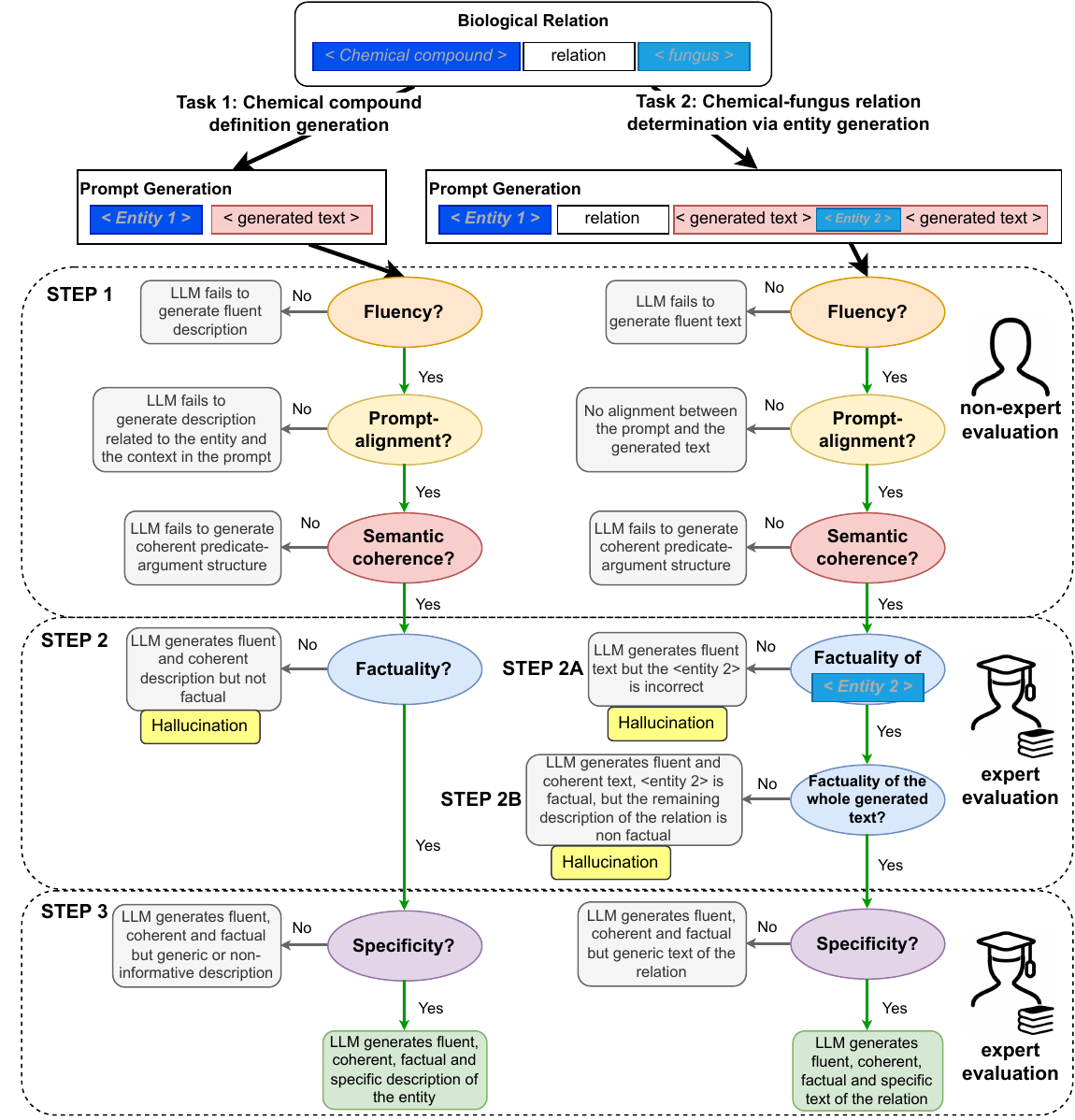}
\caption{The framework to streamline human expert evaluation of LLMs and the encoding of factual scientific knowledge in the Large Language Models, in the context of extracting biological relations. \textit{Entity 1} stands for chemical compound name. The \textit{entity 2} stands for fungus name. Fluency, prompt-alignment, and semantic coherence are assessed in \textit{STEP 1} by a non-expert (within the target domain). Then the domain expert evaluates the factuality in \textit{STEP 2} for Task 1. For Task 2, the factuality of the generated \textit{entity 2} (\textit{STEP 2A}) is verified before the entire description is assessed (\textit{STEP 2B}). Outputs that do not pass STEP 2 are classified as hallucinations. The specificity is evaluated in \textit{STEP 3}.}
\label{fig:framework}
\end{figure}

\subsection{A framework to streamline human expert evaluation of LLMs on the encoding of factual scientific knowledge}
\label{framework}
Our evaluation framework, depicted in Fig. \ref{fig:framework}, is designed to pragmatically assess the biomedical utility of LLMs by closely mirroring the human evaluation process from which it is directly derived. We propose a methodology that substantially reduces the time commitment required from experts, acknowledging the high value and limited availability of their expertise. Our approach formalises a system where non-experts, guided by specific criteria, can filter out outputs that do not require specialised biomedical knowledge, relying instead on their language proficiency. As a result, the volume of outputs that need expert evaluation is significantly reduced.

Existing human evaluation frameworks highlight distinct dimensions of this process: emphasising the impact of cognitive and utility biases on evaluation, particularly with perception-based metrics for truthfulness \cite{elangovan2024considers}; examining whether LLMs can effectively assist bioinformatics experts \cite{chen2023bioinfo}; proposing a framework for model responses along multiple dimensions, including factuality, comprehension, reasoning, potential harm, and bias \cite{singhal2023large}.

Our framework complements these by aiming to reduce the effort required from human experts when their involvement is necessary, an area which is still unexplored by existing frameworks. To the best of our knowledge, no existing framework optimises the evaluation process by composing non-expert and expert assessment phases. We argue that as LLMs are increasingly used in biomedical research, manual human evaluation will always be an integral part of model assessment.

The framework consists of three steps: 1) a non-expert evaluates whether the LLM produces text that is fluent, aligned with the prompt, and semantically coherent. If these criteria are met, the output is then passed on to an expert for further evaluation; 2) the expert verifies the factual accuracy of the output; 3) the expert assesses whether the output is generic or specific to the prompt. Each criterion is assessed qualitatively, receiving a binary score of either 0 or 1. A detailed description of the metrics is provided below.



\textbf{\textit{Fluency}} evaluates the quality of the generated text, considering two sub-criteria: syntactic correctness and style \cite{kann2018sentence,mutton2007gleu}. Syntactic correctness assesses whether a sentence conforms to grammar. Fluency is assessed first due to its simplicity, allowing non-experts to perform the evaluation effectively.
    
\textbf{\textit{Prompt-alignment}} refers to the relation of relevance and pragmatic coherence between the prompt and the generated text, whether the generated text is consistent with the input prompt and is related to the entities and context of the prompt \cite{webson2021prompt, raj2022measuring}. In other words, it checks whether the model accurately answers the question without drifting into irrelevant responses - an aspect that non-experts can effectively identify.
    
\textbf{\textit{Semantic coherence}} consists of two levels of analysis: (i) intra-sentence assesses whether the sentence is relating surface terms to expected argument types for a given predicate and (ii) inter-sentence: assessing whether adjacent sentences have a coherent discourse relation \cite{ke_ctrleval_2022}.
    
\textbf{\textit{Factuality}} is distinct from the meaning of a sentence which is conveyed by its semantic coherence. A sentence generated by a model is true if it aligns with a factual statement in the scientific literature, a database, or any other trusted source \cite{maynez-etal-2020-faithfulness}. 
    
\textbf{\textit{Specificity}} evaluates the alignment between the prompt and the answer regarding their level of abstraction. While some answers can be factually correct, they may not answer to the level of abstraction required by the prompt \cite{maynez-etal-2020-faithfulness}.


While auto-regressive language models are designed to deliver fluency, prompt-alignment and semantic coherence, text generation during decoding can lead to repetitive, incoherent or meaningless outputs \cite{holtzman_curious_2020}. As they are complementary, the fluency, prompt-alignment and semantic coherence were assessed in one step of the evaluation process (Fig. \ref{fig:framework}, \textit{STEP 1}). Moreover, the probabilistic nature of the next token prediction task also contrasts with the notion of factuality and large language models are prone to hallucination \cite{maynez_faithfulness_2020, survey_lee2023}. Hallucinations are factually incorrect statements which can be difficult to be localised by non-experts due to their expression in a fluent and semantically coherent text \cite{varshney2023stitch,curran2023hallucination}. In the context of this study, major hallucinations can be associated with the compound description or relationships with fungi, while minor hallucinations may be associated with the date of first isolation for instance. In our pragmatic approach, any output that reaches the factuality check (fulfills STEP 1) but fails to pass is considered a hallucination.

\begin{figure}[h!]
\centering
\includegraphics[width= .99\textwidth]{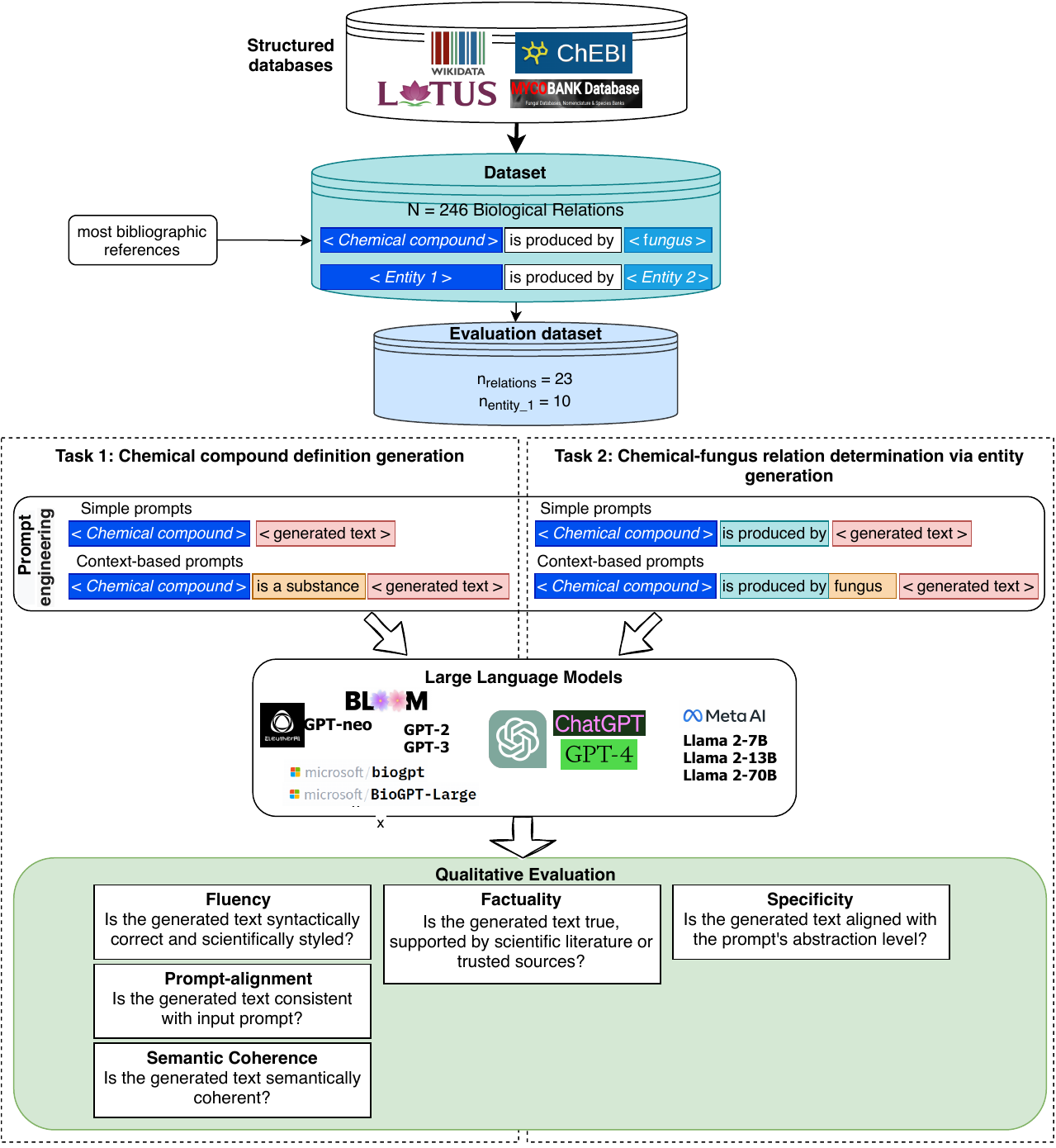}
\caption{The workflow of the performed analysis. $N$ - whole dataset with 246 selected biological relations of chemical compound-fungus pairs; $n_{relations}$ - 23 selected chemical compound-fungus pairs; $n_{entity1}$ - 10 unique chemical compounds included in those pairs; In prompt engineering: blue - \textit{entity 1}, orange - added context. Qualitative evaluation performed according the framework depicted in Fig. \ref{fig:framework}.}
\label{fig:workflow}
\end{figure}

This framework encompasses prompt definition, prompt engineering, and evaluation metrics. Then the analysis includes the extraction and curation of a dataset with biological relations, the definition of two text generation tasks for LLMs, and the selection of the target LLMs for evaluation (Fig. \ref{fig:workflow}). 
These steps allowed us to provide the first comparative analysis of LLMs in the context of tasks related to fungi and antibiotics and is transportable to other relations in bioscience, such as "A is produced by B," "A inhibits B","A activates B," etc. The code and dataset used in the study are available on \href{https://github.com/digital-ECMT/LLM_factuality_evaluation_framework.git}{GitHub}\footnote{\url{https://github.com/digital-ECMT/LLM_factuality_evaluation_framework.git}}.


\subsection{Dataset}
Investigating the biological knowledge stored in LLMs requires a collection of biological relations that are elicited and referred to in the scientific literature. Research has shown that language models (LM) exhibit bias due to imbalanced distributions in their training corpora, with performance correlated to the frequency of occurrences in the data \cite{henning2022survey, wysocki2023CL, jung2024understanding}. Scientific publications are part of the training corpus for most LLMs, thus higher prevalence in literature should lead to better performance in factual knowledge extraction. In our study, we created a dataset of biological relations accounting for the amount of supporting evidence (number of associated papers). 

Chemical compound-fungus pairs were extracted from the Wikidata Knowledge Graph using the predicate \textit{found in taxon} (\texttt{p:P703}), along with the number of bibliographic references supporting each relationship (via API). The classification and nomenclature information on the fungi was extracted from the Mycobank database\footnote{\url{https://www.mycobank.org}}. Information about antibiotic activity of chemicals were extracted from the ChEBI ontology\footnote{\url{https://www.ebi.ac.uk/chebi/downloadsForward.do}}.

A subset containing only fungal species \textbf{producing} chemicals \textbf{with} antibiotic activity was first created. The pairs with the most bibliographic references were progressively integrated until obtaining 100 distinct subject fungi, which corresponds to 123 pairs. Then, a complementary subset (with an equal number of pairs) involving \textbf{produced} chemicals but \textbf{without} antibiotic activity was added from the same list of fungi. Again, the pairs with the most bibliographic references were selected. Only molecules with a 3-STAR review in ChEBI (Chemical Entities of Biological Interest) database, indicating the highest level of manual curation and verification by ChEBI curators, were considered and only those that have an annotated role in the antimicrobial agent class (\texttt{CHEBI:33281}) were selected for positive examples of antibiotics. 
As a result, a dataset with 246 chemical compound-fungus pairs (see Fig. \ref{fig:workflow}, $N$) equally divided between chemicals with and without antibiotic activity was obtained. 
For the experiments, we randomly selected ten chemicals distributed by antibiotic activity and number of references (five chemicals with and five without antibiotic activity) (Table A.1). These ten chemicals paired with one or more fungi. In total, there were 23 chemical compound-fungus pairs (see Fig. \ref{fig:workflow}, $n_{relations}$) in the analysed subset. The selected ten chemical compounds were defined as \textit{entity 1} for two tasks. The fungi were defined as \textit{entity 2} in the second task.
Of note, the selected 23 chemical-fungus relations are described by multiple scientific publications in PubMed, all published before 2019 (for exact numbers see Table A.1). This ensures these relations are contained in the LLM training corpus, provided that the corpus contains PubMed articles or abstracts.

\subsection{Large Language Models}

In this study, we evaluated eleven LLMs (summarised in Table A.2), where six of them were released since July 2022. The models differ in terms of training corpus, vocabulary size, number of parameters, layers and maximum input sequence length. 

The final baseline set consists of: GPT-2, GPT-neo (1.3B parameters), BLOOM (3B parameters), BioGPT and BioGPT-large via huggingface API\footnote{\url{https://huggingface.co}}; Llama 2 using the llama.cpp\footnote{\url{https://github.com/ggerganov/llama.cpp}} (accessed in October 2023); all above with the default sampling procedure for text generation, i.e. greedy decoding and temperature equal 0; GPT-3 via the OpenAI API; and for ChatGPT and GPT-4 a default OpenAI user interface was used to generate outputs, each prompt in a separate session (accessed in March 2023). The maximum numbers of tokens to generate, excluding the number of tokens in the prompt is 30, except for ChatGPT, GPT-4 and Llama 2.

\label{label:methods_generative}

\subsection{Text generation tasks}

\label{label:generative_prompts}

We investigate the ability of LLMs to act as knowledge bases and establish the relation between fungi and antibiotics. Following the framework (3.1), we define two text generation tasks (Fig. \ref{fig:workflow}).

\textbf{Task 1: Chemical compound definition generation }\newline
Task 1 uses prompts for chemical compound definition generation, shown in the Table \ref{tab:text_generation_prompts} as P1-P4. For the same task we define simple prompts (P1, P2) and context-based prompts, elucidating the type of the subject entity: `a compound' or `a substance' (P3 and P4) \cite{petroni2020context}. The \textit{entity 1} represents the selected chemical compounds. We investigated ten examples of chemical compounds for the qualitative evaluation according to the previously described criteria (see section 3.1, Table A.1). Performance in Task 1 in each STEP (Fig. \ref{fig:framework}) is shown in Table \ref{tab:Text_generation_summary_percentage2}. Performance in Task 1 for optimal prompt in each STEP is shown in Table \ref{tab:task1_percentage_optimal}. 
Examples of generated answers with the qualitative evaluation are presented in Table A.3.

\textbf{Task 2: Chemical-fungus relation determination via entity generation}\newline
Task 2 uses prompts for relation determination via entity generation: \{\textit{entity 1}\}- \{relation\} - \{\textit{entity 2}\}, where \textit{entity 1} is a chemical compound name and \textit{entity 2} is a fungus name (P5-P15, Table \ref{tab:text_generation_prompts}). The prompts P10-P15 contain the context at the end, guiding the model towards a relation to a fungus, rather than to another organism capable of producing the compound. We investigated ten compounds, the same as in Task 1 described in section 3.2. Evaluation was performed according to the framework (Fig. \ref{fig:framework}) for the entire task, both with and without context. Performance in Task 2 is each STEP is shown in Table \ref{tab:Text_generation_chem-fungus_recognition_percentage2}. Performance in Task 2 for optimal prompt in each STEP is shown in Table \ref{tab:Text_generation_chem-fungus_recognition_percentage_optimal2}.

\begin{table}[b!]
\centering
\caption{Prompts used in Task 1 and Task 2.}
\label{tab:text_generation_prompts}
\resizebox{0.85\columnwidth}{!}{%
\begin{tabular}{@{}ccll@{}}
\toprule
\multicolumn{1}{l}{Task}                                                                                                             & \multicolumn{1}{l}{Type of prompt}                                               & No.  & Prompt                                                                              \\ \midrule
\multirow{7}{*}{\begin{tabular}[c]{@{}c@{}}Task 1: Chemical compound \\ definition generation\end{tabular}}                          & \multirow{3}{*}{simple prompts}                                                   & P1   & \{{\color{blue}entity 1}\}                                                                          \\
                                                                                                                                     &                                                                                   & P2   & \{{\color{blue}entity 1}\} is                                                                                \\ \cmidrule(l){2-4} 
                                                                                                                                     & \multirow{2}{*}{\begin{tabular}[c]{@{}c@{}}context-based \\ prompts\end{tabular}} & P3   & \{{\color{blue}entity 1}\} {\color{orange}is a compound}                                                            \\
                                                                                                                                     &                                                                                   & P4   & \{{\color{blue}entity 1}\} {\color{orange}is a substance}                                                            \\ \midrule
\multirow{12}{*}{\begin{tabular}[c]{@{}c@{}}Task 2: Chemical-fungus \\ relation determination \\ via entity generation\end{tabular}} & \multirow{5}{*}{simple prompts}                                                   & P5   & \{{\color{blue}entity 1}\} {\color{teal}is isolated from}                                                         \\
                                                                                                                                     &                                                                                   & P6   & \{{\color{blue}entity 1}\} {\color{teal}originally isolated from}                                                 \\
                                                                                                                                     &                                                                                   & P7  & \{{\color{blue}entity 1}\} {\color{teal}is produced by}                                                           \\
                                                                                                                                     &                                                                                   & P8  & \{{\color{blue}entity 1}\} {\color{teal}is a natural chemical compound found in}                                  \\
                                                                                                                                     &                                                                                   & P9  & \{{\color{blue}entity 1}\} {\color{teal}is a natural product found in}                                            \\ \cmidrule(l){2-4} 
                                                                                                                                     & \multirow{7}{*}{\begin{tabular}[c]{@{}c@{}}context-based \\ prompts\end{tabular}} & P10  & \{{\color{blue}entity 1}\} {\color{teal}is isolated from} {\color{orange}the fungus}                                              \\
                                                                                                                                     &                                                                                   & P11  & \{{\color{blue}entity 1}\} {\color{teal}is isolated from} {\color{orange}fungi}                                                   \\
                                                                                                                                     &                                                                                   & P12  & \{{\color{blue}entity 1}\} {\color{teal}is isolated from} {\color{orange}fungi, such as}                                          \\
                                                                                                                                     &                                                                                   & P13  & \{{\color{blue}entity 1}\} {\color{teal}is produced by} {\color{orange}the fungus}                                                \\
                                                                                                                                     &                                                                                   & P14  & \{{\color{blue}entity 1}\} {\color{teal}is produced by} {\color{orange}fungi}                                                     \\
                                                                                                                                     &                                                                                   & P15  & \{{\color{blue}entity 1}\} {\color{teal}is produced by} {\color{orange}fungi, such as}                                                 \\ \bottomrule
\end{tabular}%
}
\begin{tablenotes}\footnotesize
\item[*]
\end{tablenotes}
\end{table}

\subsection{Expert evaluation}

The outputs were evaluated and labeled according to the proposed framework by two experienced reviewers independently of each other, on each of steps 1-3. They are experts in the field, one with a PhD in microbiology and chemistry and the other with a PhD in biochemical pharmacology, both having prior experience in the manual assessment and curation of datasets.
Any discrepancies were resolved by discussion. 
Both reviewers participated in the study design and the definition of criteria. The reviewers went through a pilot process with the criteria prior to starting the study to ensure they had a similar understanding of the criteria. 
\newline

\section{Results}

\begin{figure}[htbp]
    \centering
    \begin{subfigure}[t]{0.45\textwidth}  
        \centering
        \subcaption*{A)}
        \includegraphics[width=\linewidth]{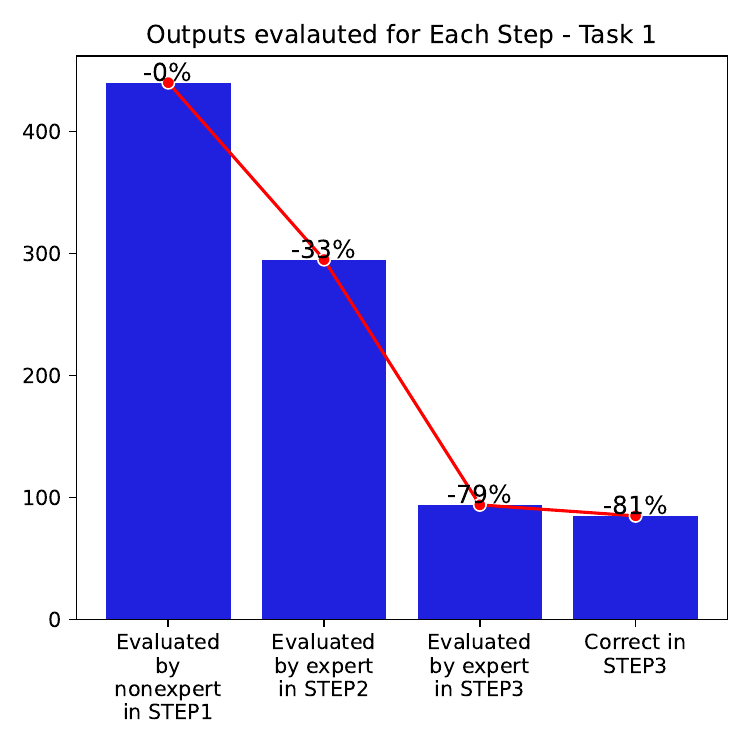}  
        \label{fig:Task_1_percentage_reduction}
    \end{subfigure}
    \hspace{0.05\textwidth}  
    \begin{subfigure}[t]{0.45\textwidth}  
        \centering
        \subcaption*{B)}
        \includegraphics[width=\linewidth]{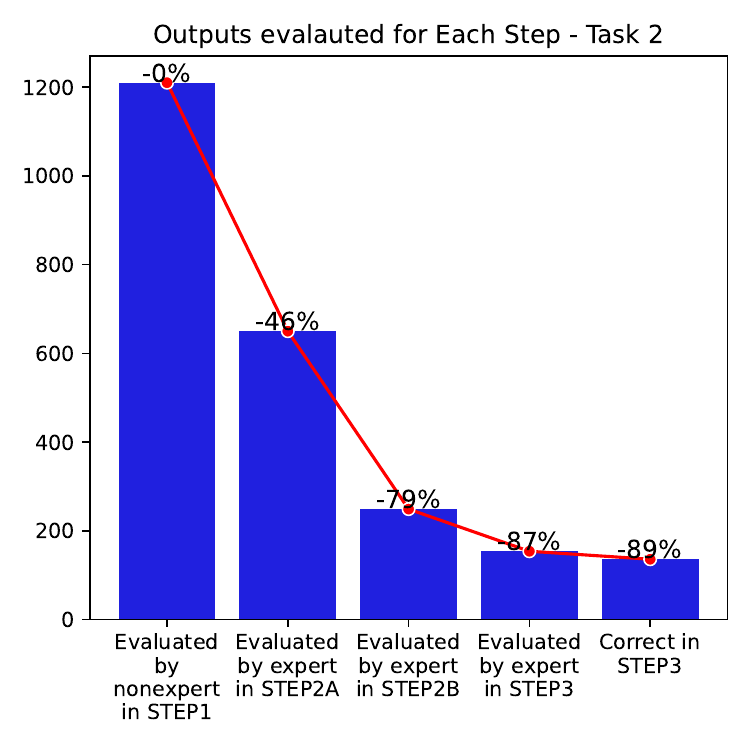}  
        \label{fig:Task_2_percentage_reduction}
    \end{subfigure}
    \caption{The number of outputs evaluated at each STEP of the framework, showing the percentage reduction compared to the initial number of outputs evaluated in STEP 1 by non-experts. Results are aggregated for 11 models and prompts from the given task: A) Task 1, B) Task 2.}
    \label{fig:Task_percentage_reduction}
\end{figure}

Evaluation of 11 models, 10 chemical compounds and 15 prompts resulted in a total of 1650 responses: 440 from Task 1 (11x10x4), 1210 from Task 2 (11x10x11).
Our framework demonstrated a significant reduction in the number of outputs requiring expert evaluation, decreasing by 33\% in Task 1 and 46\% in Task 2 (see Fig. \ref{fig:Task_percentage_reduction}), by discarding incorrect outputs in STEP 1. For Task 1, only 67\% of the outputs proceeded to the factuality check, and ultimately, only 21\% required a detailed assessment of specificity. In Task 2, 54\% of the outputs needed factuality verification for \textit{entity 2}, and subsequently, only 21\% required comprehensive factuality evaluation of the entire output. 

The responses that meet the criteria for each model and for each STEP are reported in Table \ref{tab:Text_generation_summary_percentage2} and \ref{tab:Text_generation_chem-fungus_recognition_percentage2}, and Supp.Fig. \ref{fig:P1-2}, \ref{fig:P3-4}, \ref{fig:P5-15}. 
Examples of responses generated with qualitative evaluation are presented in Table A.3 and A.4, and
optimal prompt with the best performance in Table \ref{tab:task1_percentage_optimal} and \ref{tab:Text_generation_chem-fungus_recognition_percentage_optimal2}, and Supp.Fig. \ref{fig:heatmap1}, \ref{fig:heatmap2}. In the following subsections, we provide detailed results for each individual LLM.

\subsection{Task 1: Chemical compound definition generation}
\label{label:Chemical_compound_name_recognition}

\textbf{GPT-2 and GPT-neo: no semantic coherence, no factuality.}
GPT-2 failed to generate any semantically coherent text related to the entity and the target domain (biomedicine in general).
Adding context leads to more fluent outputs but still loses semantic coherence and there is no prompt-alignment. 
GPT-neo, similarly to GPT-2, failed to generate any text or any fluent text. 
After adding context, the model asserts two compounds (Fumitremorgin C, Conidiogenone) as being produced by fungi (\textit{Fusarium sphaerospermum}, \textit{ Penicillium notatum}, respectively), however, under incorrect associations. The model failed to generate factual knowledge. 

\textbf{GPT-3: Semantically coherent answers less sensitive to prompt design. 40\% of factual answers for the best prompt (Table \ref{tab:task1_percentage_optimal}).} Hallucinates fungi names from compounds.
GPT-3 generated semantically coherent text regardless of the prompt design for all chemical compounds, both with and without context (see example of Atromentin in Table A.3). It generated a similar number of factual answers regardless of the context (30\% vs 35\%). 
GTP-3 generated incorrect fungi names (in 39.3\% of incorrect answers) that consists of the compound name in the prompt, e.g. Myriocin - Streptomyces myriocin, Chloromonicilin - Streptomyces chloromyceticus (Table A.4). 

\textbf{Bloom: no factual knowledge, low semantic coherence: biased towards treatments, repeats the same associated terms or statements.}
For prompts without context, Bloom fails to generate a syntactically correct output. 
The model tends to repeat the same words or phrases for a given entity, 
or generates names that does not exist in the literature. 
When adding the context to the prompts, Bloom is biased towards considering the compounds as a treatment for a variety of diseases. 
Frequently the answer contains repeated statements and facts, e.g. \textit{`Fumitremorgin C is a substance that is used in the treatment of cancer. It is a natural product that is used in the treatment of cancer.'}. 

\textbf{BioGPT: low specificity, adding context leads to even lower specificity, leading to a definitional generation for the compound.}
BioGPT generates fluent and syntactically correct text and tends to correspond to a more general discourse type (when contrasted to scientific discourse). 
When more context is provided, BioGPT provides a non-specific, universal definition of a compound, which is referred to regardless of the compound name (Table A.3).
After adding the prompt context `is a substance', BioGPT recognises that 5/10 compounds were produced by a fungi, but all incorrect.

\textbf{BioGPT-large: Semantically very coherent. 40\% of factual answers for the best prompt (Table \ref{tab:task1_percentage_optimal}). Hallucinations under high quality, scientific generated text are difficult to spot. }
Similarly to GPT-3, for 28\% of answers with no factuality, BioGPT-large assigned a wrong producer to the compound, e.g. \textit{'Chloromonilicin is  a new antibiotic produced by a strain of Streptomyces. It is active against Gram-positive bacteria and fungi.'} (true answer: is produced by \textit{Monilinia fructicola}). This is of particular concern, as the generated text imitates correct definitions very well, disguising the lack of factual knowledge. Such hallucination is very difficult to spot and a vigilant expert check is required.

\textbf{ChatGPT: Fluent, prompt-alignment and semantically coherent descriptions for all prompts. Hallucinations are difficult to spot. 70\% of factual answers for the best prompt (Table \ref{tab:task1_percentage_optimal}).}
Interestingly, the answers often closely resembled Wikipedia-style descriptions (7/10 compounds - Wikipedia, 2/10 - Wikidata). For prompts with no context, the statements were longer and divided into paragraphs (mostly three; note, that a limit was not set on the length of generated answer for ChatGPT). 
Comparing to BioGPT-large, ChatGPT produces even more fluent, prompt-aligned and semantically coherent answers. Due to its semantic fluency, evaluation of the factual knowledge requires both expert knowledge and the time-consuming fact verification. In other words, generated answers can be easily regarded as true and factual for a non-expert, creating a real risk of deriving false facts from the model.

\textbf{GPT-4: Top performer from the list and epistemically-aware (admits to lack of knowledge). 80\% of factual answers for the best prompt (Table \ref{tab:task1_percentage_optimal}).}
The model did not guess the definition if it did not recognise the name. For two chemical compounds, the model generated a request for more context or referred to the knowledge cutoff date of 2021-09.
In general, adding context did not improve neither the factuality nor specificity metrics, which remained at 70\% (for two prompts in total, P1-P2 vs P3-P4). The quoted names of the fungus that produces the chemical were incorrect. In general, the model generated fluent, semantically coherent, highly specific text (Table A.4). We observed less hallucinations compared to ChatGPT.

\textbf{Llama 2: Exhibits a proclivity for generating inaccurate comparisons, particularly evident in chemical contexts.}
Llama 2-70B model produced 20\% (4/20) responses that were both specific and factual  (20\% of factual answers for each prompt, P1 and P2). Smaller versions of Llama2 demonstrated inferior performance. Adding context failed to enhance either factuality or specificity metrics; instead, such augmentation resulted in a noteworthy decline, approaching zero, across all tested Llama 2 models. The Llama2 models displayed a noticeable tendency to generate comparative phrases, particularly evident when tasked with chemical comparisons, frequently resorting to statements like `100 times more toxic than'.
In summary, the Llama 2 models failed to generate text characterized by fluency, semantic coherence, and high specificity.

\begin{table}[]
\centering
\caption{Performance in Task1: \textit{chemical compound definition generation}. Each step from the framework (see Section \ref{framework}) evaluated separately. X/20 represents the number of correct responses versus the total number of outputs for a given type of prompt (each prompt has 10 outputs). The best results are highlighted in bold.}
\label{tab:Text_generation_summary_percentage2}
\resizebox{0.6\columnwidth}{!}{%
\begin{tabular}{@{}lllllll@{}}
\toprule
             & \multicolumn{6}{c}{Task 1: Chemical compound definition generation}                        \\ \cmidrule(l){2-7} 
             & \multicolumn{3}{c}{simple prompts P1-P2} & \multicolumn{3}{c}{context-based prompts P3-P4} \\ \cmidrule(l){2-7} 
Model        & STEP 1       & STEP 2      & STEP 3      & STEP 1          & STEP 2        & STEP 3        \\ \midrule
GPT-2        & 2/20    & 0/20          & 0/20            & 9/20        & 0/20              & 0/20              \\
GPT-3        & 20/20   & 6/20    & 6/20    & 20/20      & 7/20      & 7/20      \\
GPT-neo      & 8/20     & 0/20            & 0/20            & 11/20       & 0/20              & 0/20              \\
Bloom        & 4/20     & 0/20            & 0/20            & 6/20        & 0/20              & 0/20              \\
BioGPT       & 14/20    & 1/20     & 0/20            & 19/20       & 7/20      & 2/20      \\
BioGPT-large & 16/20    & 6/20    & 5/20    & 20/20      & 7/20      & 6/20      \\
ChatGPT      & 20/20   & 12/20   & 12/20   & 20/20      & 11/20     & 10/20     \\
GPT-4        & \textbf{20/20}   & \textbf{14/20}   & \textbf{14/20}   & \textbf{20/20}      & \textbf{14/20}     & \textbf{14/20}     \\
Llama 2-7B   & 16/20    & 2/20    & 2/20    & 4/20        & 0/20              & 0/20              \\
Llama 2-13B  & 12/20    & 2/20    & 2/20    & 9/20        & 0/20              & 0/20              \\
Llama 2-70B  & 13/20    & 4/20    & 4/20    & 12/20       & 1/20       & 1/20       \\ \bottomrule
\end{tabular}%
}
\begin{tablenotes}\footnotesize
\item[*]
\end{tablenotes}
\end{table}

\begin{table}[]
\centering
\caption{Performance for optimal prompts in Task 1: \textit{chemical compound definition generation}. For each model an optimal prompt (highest score in STEP 2) was selected. Evaluation according to the framework (Fig. \ref{fig:framework}). X/10 is the number of correct responses vs total number of chemicals.}
\label{tab:task1_percentage_optimal}
\resizebox{.55\columnwidth}{!}{%
\begin{tabular}{@{}lllll@{}}
\toprule
Model                                     & Optimal Prompt          & STEP1                                                                                       & STEP2                                                                                              & STEP 3                                         \\ \midrule
GPT-3                                      & \{entity 1\} is a substance ...            & 10/10                                                                                       & 4/10                                                                                               & 4/10                                         \\ \midrule
BioGPT                                      & \{entity 1\} is a compound ...            & 9/10                                                                                       & 5/10                                                                                               & 1/10                                         \\ \midrule
BioGPT-large                                      & \{entity 1\} is a compound ...            & 10/10                                                                                       & 4/10                                                                                               & 4/10                                         \\ \midrule
ChatGPT                                    & \{entity 1\} is a substance ...             & 10/10                                                                                      & 7/10                                                                                               & 6/10                                         \\ \midrule
GPT-4                     & \{entity 1\} ...               & 10/10                                                                                       & 8/10                                                                                               & 8/10                                         \\ \midrule
Llama 2-7B                                 & \{entity 1\} ...        & 9/10                                                                                       & 2/10                                                                                                  & 2/10                                            \\ \midrule
\multirow{2}{*}{Llama 2-13B}                    & \{entity 1\} ...        & 7/10                                                                                      & 1/10                                                                                               & 1/10                                         \\ 
& \{entity 1\} is ...        & 5/10                                                                                      & 1/10                                                                                               & 1/10                                         \\ \midrule
\multirow{2}{*}{Llama 2-70B}                                & \{entity 1\} ...        & 7/10                                                                                       & 2/10                                                                                               & 2/10                                         \\  
& \{entity 1\} is ...        & 6/10                                                                                       & 2/10                                                                                               & 2/10                                         \\ \bottomrule
\end{tabular}%
}
\end{table}

\subsection{Task 2: Chemical-fungus relation determination via entity generation}

\textbf{GPT-3 produces at least one factual relation for 6/10 compounds.}
GPT-3 generated 51\% (56/110) answers containing a fungus name, where 23.6\% (26/110) were factual compound-fungus relations, but only in 13.6\% (15/110) the whole relation description was factual (STEP 2B). 
Alternariol turned out to be a chemical with the most factual relations (8 out of 10 tested prompts), which aligns with the results from section 3.1 for GPT-3 (Table A.5).

\textbf{BioGPT biased towards \textit{Aspergillus}, limited proportion of factual relations.}
BioGPT generated factual relations with fungi names for only 13.6\% (15/110) of the cases, of which only ten answers had factual knowledge. 
Out of 62 generated answers that contained a fungus, 45 were \textit{Aspergillus}, of which only 11 answers were factual. 

\textbf{BioGPT-large: at least one factual chemical-fungus relation for 9/10 compounds.}
This implies that it is possible to achieve 90\% of factual relations if picking the right prompt for each chemical, showing the dependency on prompt optimisation. The model gave the most factual answers for Ergosterol (8 out of 10 tested prompts). 
Contrary to previous models, we observed that BioGPT-large did not generate additional context, when unclear (e.g. the names of the discoverers or the year of discovery of the compound if not known) reducing the overall number of hallucinations.

\textbf{ChatGPT: coherent answers, regardless of the prompt, but only 31.8\% (35/110) of them were fully factual for the whole generated description (STEP 2B).}
The lack of factuality was mainly due to the incorrect statement of the year of discovery/isolation of a given compound, the name of the discoverer or the class of the compound.
The model provided all 11 fungal-name responses for eight compounds. The exception was one prompt for Conidiogenone and seven prompts for Ergosterol. The rest of the answers given for Ergosterol (4) were 100\% factual with fungus relation. 
For three compounds all answers were factual (100\% for Fumitremorgin C, Alternariol, Verrucarin A (Table A.6)) and for four compounds all prompts led to incorrect relations. 

\textbf{GPT-4 gives either the factual chemical-fungus relations or no answer at all.}
For four compounds, all answers were factual (11/11, 100\% for Fumitremorgin C, Alternariol, Verrucarin A, Myriocin). The model for all prompts for Ergosterol did not generate any relations with the fungus and factually recognized the compound by citing its characteristics.
In the case of 13.6\% (15/110) responses, the model asked for more context and/or indicated that the question is incomplete. 
The hallucination (lack of factual knowledge) was observed for ~56\% of outputs, mainly as an incorrect statement of the year of discovery/isolation of a given compound or the incorrect citation of the name of the synonym of the fungus.

\textbf{Llama 2: the factuality increases with the model size.}
As the model size increases, the Llama 2 model exhibits an improved capacity to generate fungal names and factual chemical compound-fungus relations, constituting 28.2\% (31/110) of correct relations for Llama 2-70B. Notably, the Llama 2 70B accurately generated complete definitions, inclusive of the fungus name, with a success rate of 12.7\% (14/110).
Irrespective of the model size, Llama 2 exhibited a proclivity for response repetition and looping.
Concerning the chemical compound Verrucarin A, all Llama 2 models erroneously postulated the name of a fungus (Verrucaria) unrelated to the actual producer of this compound (\textit{Albifimbria verrucaria} or \textit{Myrothecium verrucaria}), with instances recorded at 36.4\% (4/11) for Llama 2-7B, 45.5\% (5/11) for Llama 2-13B, and 36.4\% (4/11) for Llama 2-70B.

\begin{table}[tb]
\centering
\caption{Performance in Task 2: \textit{chemical-fungus relation determination via entity generation}. Each step from the framework (see Section \ref{framework}) evaluated separately. X/110 represents the number of correct responses versus the total number of responses for a given model, with 110 derived from 11 prompts each applied to 10 chemicals. The best results are highlighted in bold.}
\label{tab:Text_generation_chem-fungus_recognition_percentage2}
\resizebox{.7\columnwidth}{!}{%
\begin{tabular}{@{}llllll@{}}
\toprule
             & \multicolumn{5}{c}{Task 2: Chemical-fungus relation determination via entity generation}                                                                                                                                                                                                                                                         \\ \cmidrule(l){2-6} 
             & \multicolumn{5}{c}{simple prompts + context-based prompts P5-P15}                                                                                                                                                                                                                                                                                \\ \cmidrule(l){2-6} 
             & \multicolumn{1}{c}{\multirow{2}{*}{STEP 1}} & \multicolumn{2}{c}{STEP 2A}                                                                                                                                                                           & \multicolumn{1}{c}{\multirow{2}{*}{STEP 2B}} & \multicolumn{1}{c}{\multirow{2}{*}{STEP 3}} \\ \cmidrule(lr){3-4}
Model        & \multicolumn{1}{c}{}                        & \begin{tabular}[c]{@{}l@{}}occurrence of \\ fungus name in \\ the generated text\end{tabular} & \begin{tabular}[c]{@{}l@{}}factual occurrence \\ of fungus name in \\ the generated text\end{tabular} & \multicolumn{1}{c}{}                         & \multicolumn{1}{c}{}                        \\ \midrule
GPT-2        & 0/110                                             & 0/110                                                                                          & 0/110                                                                                                  & 0/110                                         & 0/110                                             \\
GPT-3        & 57/110                                             & 56/110                                                                                    & 26/110                                                                                          & 15/110                                 & 15/110                                              \\
GPT-neo      & 29/110                                             & 28/110                                                                                  & 4/110                                                                                            & 2/110                                   & 0/110                                             \\
Bloom        & 13/110                                             & 12/110                                                                                    & 3/110                                                                                            & 2/110                                   & 2/110                                             \\
BioGPT       & 62/110                                             & 57/110                                                                                  & 15/110                                                                                          & 10/110                                  & 2/110                                             \\
BioGPT-large & 84/110                                           & 74/110                                                                                  & 44/110                                                                                            & 18/110                                 & 14/110                                             \\
ChatGPT      & 103/110                                             & 101/110                                                                                 & 49/110                                                                                          & 35/110                                 & 35/110                                             \\
GPT-4        & \textbf{87/110}                                             & \textbf{77/110}                                                                                    & \textbf{56/110}                                                                                          & \textbf{48/110}                                     & \textbf{48/110}                                             \\
Llama 2-7B   & 69/110                                & 36/110                                                                                  & 7/110                                                                                            & 2/110                                   & 2/110                                  \\
Llama 2-13B  & 70/110                                & 49/110                                                                                  & 14/110                                                                                          & 8/110                                   & 4/110                                  \\
Llama 2-70B  & 76/110                                & 62/110                                                                                  & 31/110                                                                                          & 14/110                                 & 14/110                                \\ \bottomrule
\end{tabular}%
}
\begin{tablenotes}\footnotesize
\item[*]
\end{tablenotes}
\end{table}

\begin{table}[]
\caption{Performance for optimal prompts in Task 2: \textit{chemical-fungus relation determination via entity generation}. For each model an optimal prompt (highest score in STEP 2B) was selected. Evaluation according to the framework (Fig. \ref{fig:framework}). X/10 is the number of correct responses vs total number of chemicals.}
\label{tab:Text_generation_chem-fungus_recognition_percentage_optimal2}
\resizebox{\columnwidth}{!}{%
\begin{tabular}{@{}lllll@{}}
\toprule
\multicolumn{1}{c}{\multirow{2}{*}{Model}} & \multicolumn{1}{c}{\multirow{2}{*}{Optimal Prompt}} & \multicolumn{2}{c}{STEP 2A}                                                                                                                                                                     & \multicolumn{1}{c}{\multirow{2}{*}{STEP 2B}} \\ \cmidrule(lr){3-4}
\multicolumn{1}{c}{}                       & \multicolumn{1}{c}{}                                & \begin{tabular}[c]{@{}l@{}}occurrence of fungus name \\ in the generated text\end{tabular} & \begin{tabular}[c]{@{}l@{}}factual occurrence of fungus \\ name in the generated text\end{tabular} & \multicolumn{1}{c}{}                         \\ \midrule
GPT-3                                      & \{entity 1\} is produced by the fungus ...            & 8/10                                                                                       & 5/10                                                                                               & 4/10                                         \\ \midrule
GPT-neo                                    & \{entity 1\} is produced by the fungus ...            & 5/10                                                                                       & 2/10                                                                                               & 0/10                                            \\ \midrule
Bloom                                      & \{entity 1\} is produced by the fungus ...            & 7/10                                                                                       & 2/10                                                                                               & 1/10                                         \\ \midrule
\multirow{2}{*}{BioGPT}                    & \{entity 1\} is produced by fungi, such as ...        & 10/10                                                                                      & 3/10                                                                                               & 1/10                                         \\
                                           & \{entity 1\} is isolated from fungi, such as ...      & 10/10                                                                                      & 3/10                                                                                               & 2/10                                         \\ \midrule
\multirow{2}{*}{BioGPT-large}              & \{entity 1\} is produced by the fungus ...            & 10/10                                                                                      & 6/10                                                                                               & 2/10                                         \\
                                           & \{entity 1\} is isolated from the fungus ...          & 10/10                                                                                      & 6/10                                                                                               & 3/10                                         \\ \midrule
ChatGPT                                    & \{entity 1\} originally isolated from ...             & 10/10                                                                                      & 6/10                                                                                               & 1/10                                         \\ \midrule
\multirow{2}{*}{GPT-4}                     & \{entity 1\} is isolated from fungi ...               & 9/10                                                                                       & 6/10                                                                                               & 5/10                                         \\
                                           & \{entity 1\} is isolated from fungi, such as ...      & 9/10                                                                                       & 6/10                                                                                               & 5/10                                         \\ \midrule
Llama 2-7B                                 & \{entity 1\} is produced by fungi, such as ...        & 9/10                                                                                       & 0/10                                                                                                  & 0/10                                            \\ \midrule
Llama 2-13B                                & \{entity 1\} is produced by fungi, such as ...        & 9/10                                                                                       & 2/10                                                                                               & 1/10                                         \\ \midrule
Llama 2-70B                                & \{entity 1\} is produced by fungi, such as ...        & 9/10                                                                                       & 5/10                                                                                               & 2/10                                         \\ \bottomrule
\end{tabular}%
}
\end{table}

\section{Discussion}

Our observations, initially rooted in the analysis of fungus-chemical interactions, hold relevance for a range of specific domains within the biomedical field. In this section, we detail these broadly applicable insights, directing the evaluator's attention to essential concerns. Key observations include biases in LLM outputs towards more prevalent topics, difficulties in accurately identifying rare entities, a tendency for outputs to stray from the intended biomedical context, and significant variability in performance influenced by the design of the prompts. 

\subsection{LLMs' biases towards compounds and fungi}

\subsubsection{Correctly generated relations are homogeneous and prevalent.}
Considering the models with outputs that contain the fungus name, in majority they recognise one specific factual chemical-fungus relation, although the chemical can be produced by several fungi (RQ1). For instance, in all generated relations which were factual, for Fumitremorgin C the fungus was either \textit{Aspergillus fumigatus} or \textit{Aspergillus} (39 out of 39 correct fungi name, from 66 generated fungi in total). 
Factually generated relations for Ergosterol are with two fungi: \textit{Aspergillus} (11/23) and \textit{Ganoderma lucidum} (7/23). Interestingly, \textit{Aspergillus} is output by BioGPT and \textit{Ganoderma lucidum} by BioGPT-Large. Although trained on the same corpus, we hypothesise that the higher number of parameters in BioGPT-Large (347M vs 1.5B) led to a more specific relation (RQ2).

\subsubsection{Bias towards \textit{Aspergillus} - the most cited fungus in PubMed.}
We observe that the models are biased towards certain compounds (RQ1). For instance, BioGPT generates the fungus \textit{Aspergillus} as the answer in most of the prompts (79\%, 45/57), which is rarely correct (only 11 times; 19\%) (Table A.7). We attribute this bias to the imbalance in the training corpus, as the \textit{Aspergillus} is a ubiquitous fungus in home and hospital environments. 
The PubMed search outputs almost 60k references related to \textit{Aspergillus}. The most cited compound from our analysis \textit{Ergosterol} outputs $\sim$6000, second \textit{Aphidicolin} $\sim$2500. Thus, we argue \textit{Aspergillus} is over-represented in the training corpus and occurs as a statistically most probable fungus to be related to chemicals.

\subsubsection{Rare chemical compound confused with an overrepresented biological process.}
Another example, \textit{Conidiogenone} was consistently not recognised by the best performing models (GPT-4, ChatGPT, BioGPT-large). We argue that the models confuse Conidiogenone with \textit{conidiogenesis} or \textit{conidiation} (RQ1). \textit{Conidiogenesis} is a widespread morphogenetic process that filamentous fungi undertake for dispersion. \textit{Conidiation} is also important for pathogenicity of phytopathogens. Understanding the cellular mechanism of \textit{conidiation} is a highly relevant research topic, thus it has been extensively studied. \textit{Conidiogenone} is associated with nine results in a PubMed search, \textit{conidiogenesis} with 346 and \textit{conidiation} with 6047. The strongest and widespread stimulus for \textit{conidiation} among filamentous fungi is the exposure of hyphae to the air. Often the generated texts focused on this aspect of the process, suggesting that the model encoded the signal for the representation of the process instead of the chemical compound.  

\subsubsection{Output not focused on fungi despite the prompt context.}
In the case of Ergosterol, which is a compound widely known for its significance in the scientific community, the LLM specifically trained for the bio-domain (BioGPT-Large) outperforms the rest (RQ3). It generated seven factual Ergosterol-fungus relations from 11 prompts (Table A.8). For the other models, the relation to fungus was either minimally addressed or not considered at all, particularly for the models that have generally demonstrated superior efficacy (i.e., GPT-4 with 0/11 and ChatGPT with 4/11 factual responses). We observe that GPT-4 and ChatGPT focus on Ergosterol's biosynthesis in fungi and relevance as an antifungal target, rather than on the producing fungi, despite the prompt that explicitly defines such context. Importantly, these two models do not generate text as a continuation of the prompt, but they generate new paragraphs.

\subsection{Limited ability of LLMs as Knowledge Bases}

In this study, we provide evidence of the limitations in the determination of factual knowledge from LLMs in the text generation tasks. Best models were able to provide only 6/10 of the factual description of chemical entities, and only 5/10 of the factual description of chemical-fungus relations. Despite small sample size in our evaluation, it clearly shows that the performance is insufficient for a systematic and reliable application (RQ1, RQ4). 

\subsection{High dependency and variability due to prompt design choices}

A desirable property of a LLM is to a consistent answer regardless of the prompts design. For instance, the prompt `\{{\color{blue}entity 1}\} is produced by fungi, such as' should be equivalent to `\{{\color{blue}entity 1}\} is isolated from fungi, such as' because it refers to the same biological relation. However, we observe a variation in the output as the evidence of sensitivity for prompt design (RQ1). Similarly to Masked Language Models \cite{zhou2023large}, in order to fully exploit GPTs capabilities, prompt engineering is required. In our study, we observe that providing context in the prompt improves the performance. Of note, ChatGPT and GPT-4 are the least sensitive and produce new paragraphs instead of finishing the prompt (RQ3).

\subsection{Llama 2, despite large size, responses lack factuality and specificity}

The analysis of Llama 2's capabilities shows that factuality improves with increased model size. An example of this correlation is Ergosterol, not recognised in Task1 by 7B and 13B versions, but correctly defined by Llama 2 70B. Yet, even the biggest Llama2 variant, did not match the factual strength of GPT-4, ChatGPT, or BioGPT-large, delivering longer, less fluent, and semantically weaker responses. In generating chemical definitions, GPT models surpassed Llama 2 with more nuanced outputs.

\subsection{Best overall performance was on large corpora, large parameters and most extensive human feedback alignment}

GPT-4 outperforms ChatGPT (as well the rest tested models) in terms of factuality of the answer (RQ3). It scores 70\% (7/10) and 43.6\% (48/110) for chemical entity recognition and chemical-fungi relation recognition, respectively. In general, the factual knowledge of the entire generated text was higher (still significantly higher in GPT-4), but the output did not contain the relation that was prompted for, so they were considered not specific. 
The hallucinations in GPT-4 are the lowest (54.4\%). Importantly, GPT-4 produced only names of fungi which actually produce chemical compounds.

\subsection{Limitations}
We recognise the following limitations of this investigation. First, LLMs are evaluated on a specific biological dataset related to chemical compounds, fungi and antibiotic activity. The models can perform differently for other biological relations. The sample size of 10 compounds is relatively small and increasing the size would lead to more representative results. 

Although we evaluated 15 prompt designs, the prompt engineering could be improved with e.g. a systematic search or optimal prompt algorithms. No fine-tuning, nor few-shot learning was performed. 
The analysis used of-the-shelf LLMs, focusing on their factuality in a specific scientific task. Despite the models' varying corpora, parameters, and design intents, the study did not aim to equalise these factors but rather to explore which features contribute to factual knowledge-based inference in LLMs.

A total of $>$ 150 000 fungi species have been described in the literature \cite{WangFungi2022}. The naming of fungal species is subject to change over time \cite{WangFungi2022, RichardsFungi2017}, and moreover, the number of fungal names continue to increase over time. Despite the improvements in information sharing, the changing fungal nomenclature rules \cite{aime2021publish, turland2018international} and providing a tool for name standardisation, the existing literature still contains plenty of synonyms, homonyms, orthographic variants and misused names that are not in accordance with the standard nomenclature system \cite{lucking2020unambiguous}. In addition, some taxon names in the databases may be variants, synonyms or invalidly published names for various reasons. This may lead to inaccurate search results and renders the assessment of factual knowledge in generated relations challenging even for a domain expert.

\section{Conclusions}

This work explores the potential of LLMs to encode domain-specific factual knowledge, with an emphasis on the biomedical field. To assess this, we introduced a framework to streamline the labor-intensive process of human expert evaluation, using non-experts for preliminary assessments to optimise the time and expertise of specialists, without sacrificing accuracy and efficiency. In our study, this approach resulted in a 33\% and 46\% reduction in experts' effort for two different tasks.
The framework replicates the process of human evaluation, which remains an indispensable step regardless of the presence of automated evaluation systems \cite{Bavaresco2024}. This is due to the fact that every user in the biomedical field will invariably wish to conduct manual checks of the model using a variety of familiar examples. Essentially, this framework serves as a directive on the aspects to focus on. 

The results reveal that, although recent advancements have enhanced fluency and semantic coherence in LLMs, a reliable performance of the models to operate as a domain-specific knowledge base remains elusive. The factuality of the generated content, especially in biological contexts, is still lacking, with models showing a pronounced bias towards entities that are more prevalent in the training corpus. Our findings align with previous studies across various domains \cite{razeghiImpactPretrainingTerm2022, wangRetrieveWhatYou, zhaoCalibrateUseImproving2021, kassnerArePretrainedLanguage2020, kandpalLargeLanguageModels2023}, confirming co-occurrence biases irrespective of model size. 
Notably, GPT-4, while outperforming others by demonstrating mechanisms of epistemic awareness, acknowledging its knowledge limitations, thereby preventing certain hallucinations. It recognised 70\% of chemical compounds and produced less than half of the factual and specific relations to fungi. This compares to the Llama 2 model, which recognized 20\% of compounds and 12.7\% of relations to fungi. Our study addresses the increasing demand for thorough assessments of LLMs in various fields, advancing beyond the standard benchmarks that are commonly employed. This research complements existing works investigating LLM performance within the biomedical domain, specifically probing whether these models simply recall information or truly understand the symbolic connections between chemicals and fungi when generating definitions.

\textbf{Future work.} As a continuation of this study, we plan to further assess LLMs for multi-hop biomedical inference. We plan to extend our proposed framework addressing the need for efficient qualitative and quantitative evaluation of responses generated by LLMs. Such frameworks should go beyond the well-known benchmarks and be transferable to other biomedical domains, such as the one exemplified in this study (chemical-fungus relation). A promising future direction is the use of robust retrieval augmented generation (RAG) mechanisms  integration with robust, high-recall retrieval augmentation mechanisms, 
e.g. \cite{delmasRelationExtractionUnderexplored2023,wysocki2024llm} in coordination with symbolic reasoning methods.

\section*{CRediT authorship contribution statement}

\textbf{Magdalena Wysocka:} Writing – review and editing, Writing – original draft, Software, Methodology, Formal analysis, Data curation, Visualization, Conceptualization. \textbf{Oskar Wysocki:} Supervision, Writing – review and editing, Formal analysis, Methodology, Conceptualization. \textbf{Maxime Delmas:} Writing – review and editing, Writing – original draft, Software. \textbf{Vincent Mutel:} Project administration, Methodology. \textbf{Andr\'{e} Freitas:} Writing – review and editing, Supervision, Investigation.

\section*{Declaration of competing interest}

The authors declare that they have no known competing financial interests or personal relationships that could have appeared to influence the work reported in this paper.

\section*{Acknowledgments}

J. Dumoulin, J. Rossier and C. Verzat for their support.

\section*{Funding}

This project has received funding from the European Union's Horizon 2020 research and innovation programme under grant agreement No 965397. This work was supported by the IDIAP Research Institute and has been done in collaboration with the company Inflamalps SA and is supported by the Ark Foundation. The funding bodies played no role in the design of the study, research, writing and publication of the paper.

\bibliographystyle{unsrtnat}
\bibliography{custom}






\appendix

\section{Supplementary data}
The following is the Supplementary material related to this article.

\newcommand{\supplementaryfigures}{
    \setcounter{figure}{0}  
    \renewcommand{\thefigure}{A.\arabic{figure}}  
    \renewcommand{\figurename}{Figure}  
}

\supplementaryfigures  







\begin{figure}[h!]
\centering
\includegraphics[width= .85\textwidth]{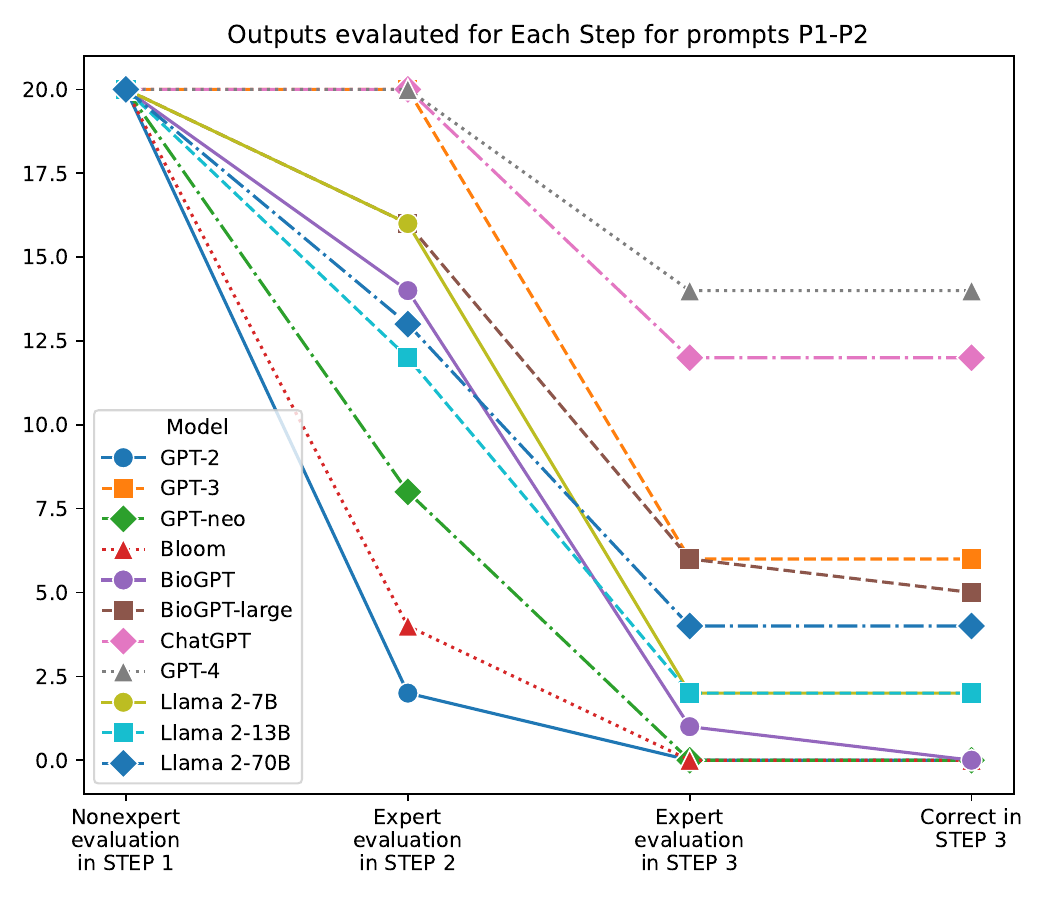}
\caption{The number of outputs evaluated at each STEP of the framework in Task 1 for simple prompts, showing the percentage of results evaluated by non-expert and expert. Results are aggregated for prompts P1-P2 from the given model.}
\label{fig:P1-2}
\end{figure}

\begin{figure}[h!]
\centering
\includegraphics[width= .85\textwidth]{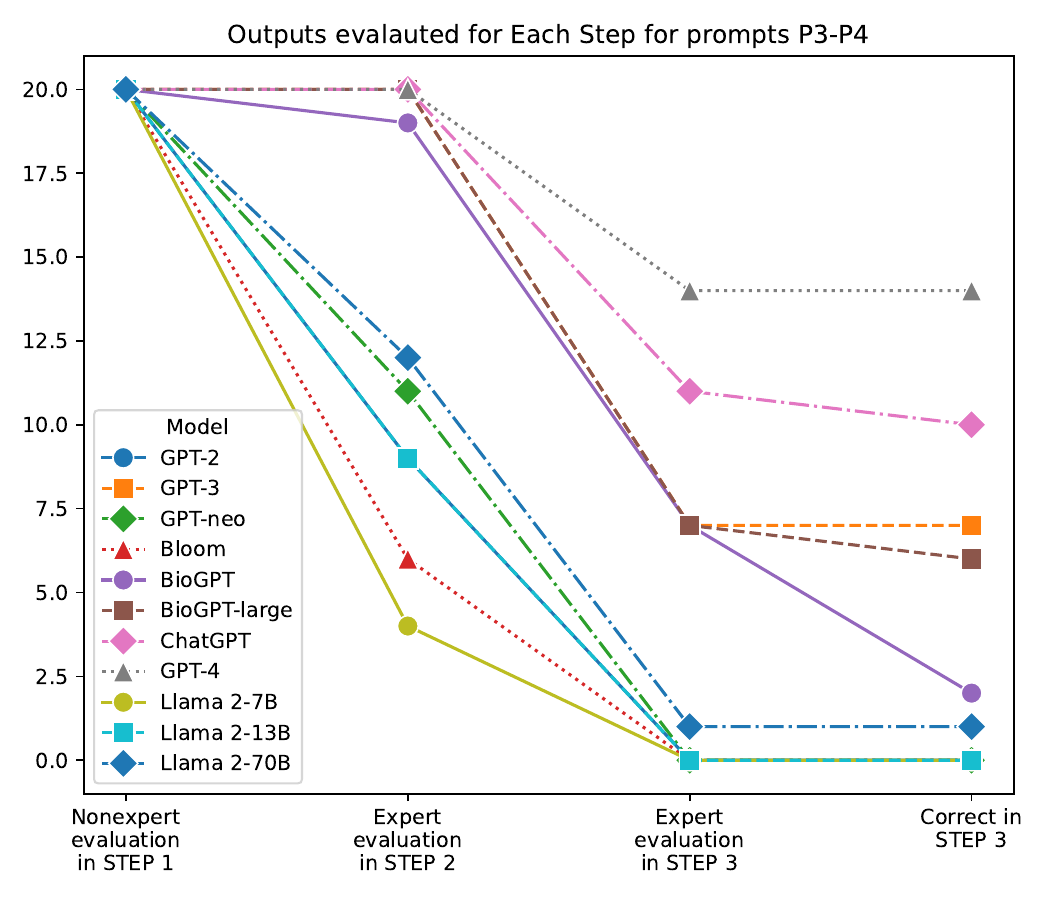}
\caption{The number of outputs evaluated at each STEP of the framework in Task 1 for context-based prompts, showing the percentage of results evaluated by non-expert and expert. Results are aggregated for prompts P3-P4 from the given model.}
\label{fig:P3-4}
\end{figure}

\begin{figure}[h!]
\centering
\includegraphics[width= .85\textwidth]{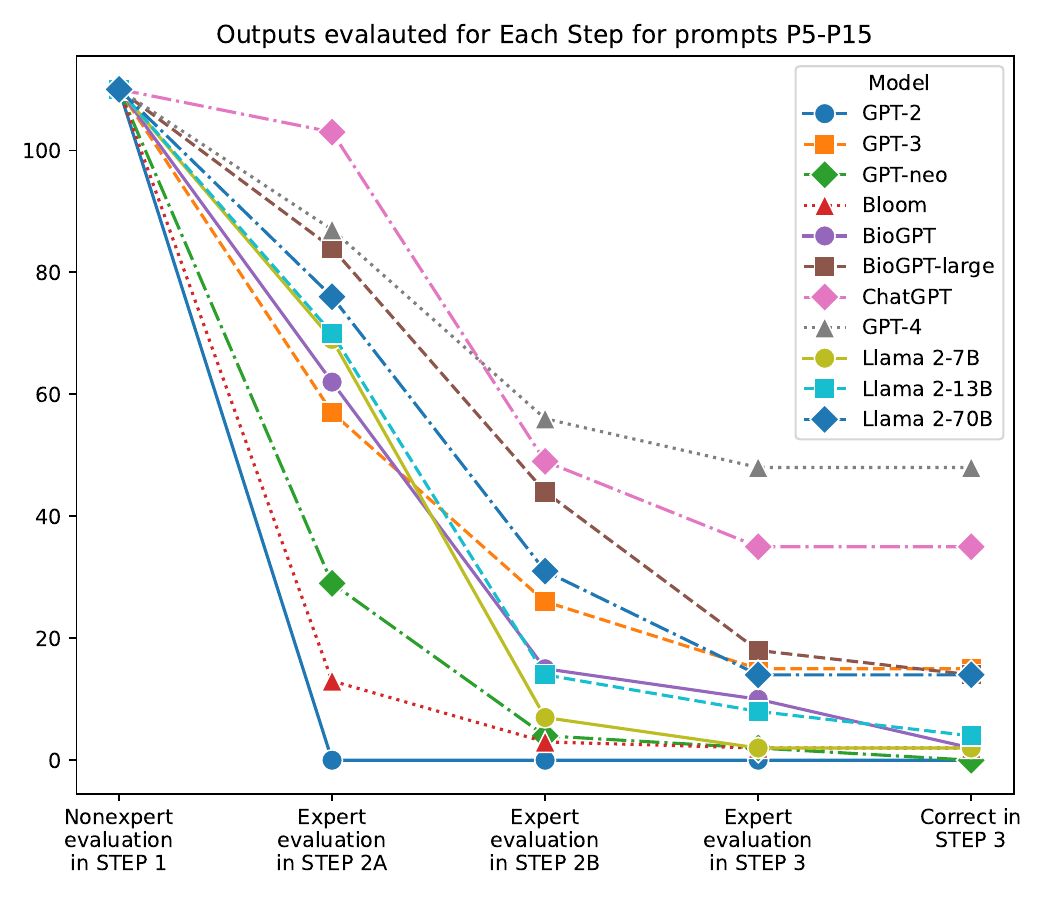}
\caption{The number of outputs evaluated at each STEP of the framework in Task 2, showing the percentage of results evaluated by non-expert and expert. Results are aggregated for all prompts P5-P15 from the given model.}
\label{fig:P5-15}
\end{figure}

\begin{figure}[htbp]
    \centering
    \begin{subfigure}[t]{0.38\textwidth}  
        \centering
        \subcaption*{A)}
        \includegraphics[width=\linewidth]{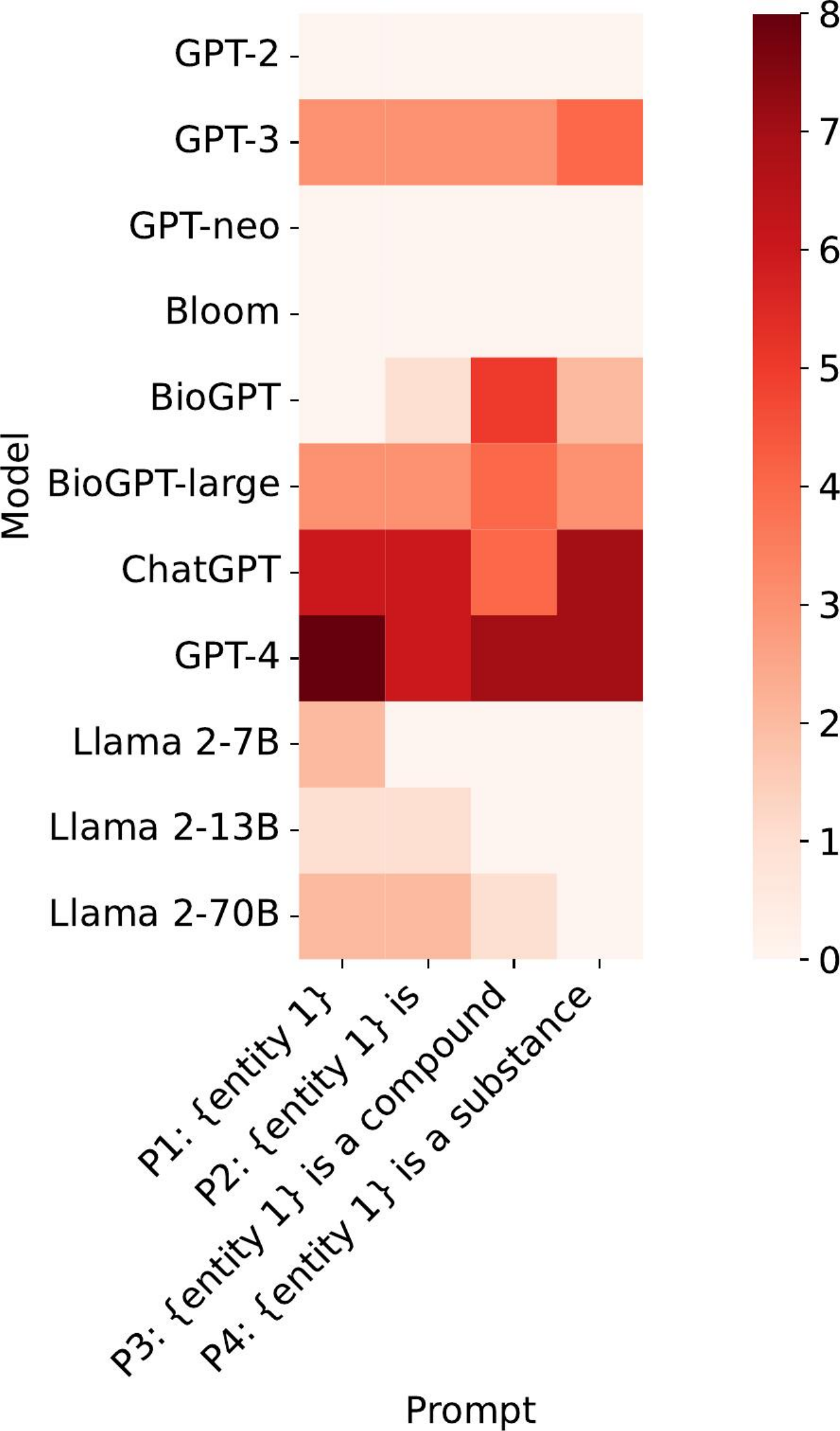}  
        \label{fig:sub1}
    \end{subfigure}
    \hspace{0.05\textwidth}  
    \begin{subfigure}[t]{0.45\textwidth}  
        \centering
        \subcaption*{B)}
        \includegraphics[width=\linewidth]{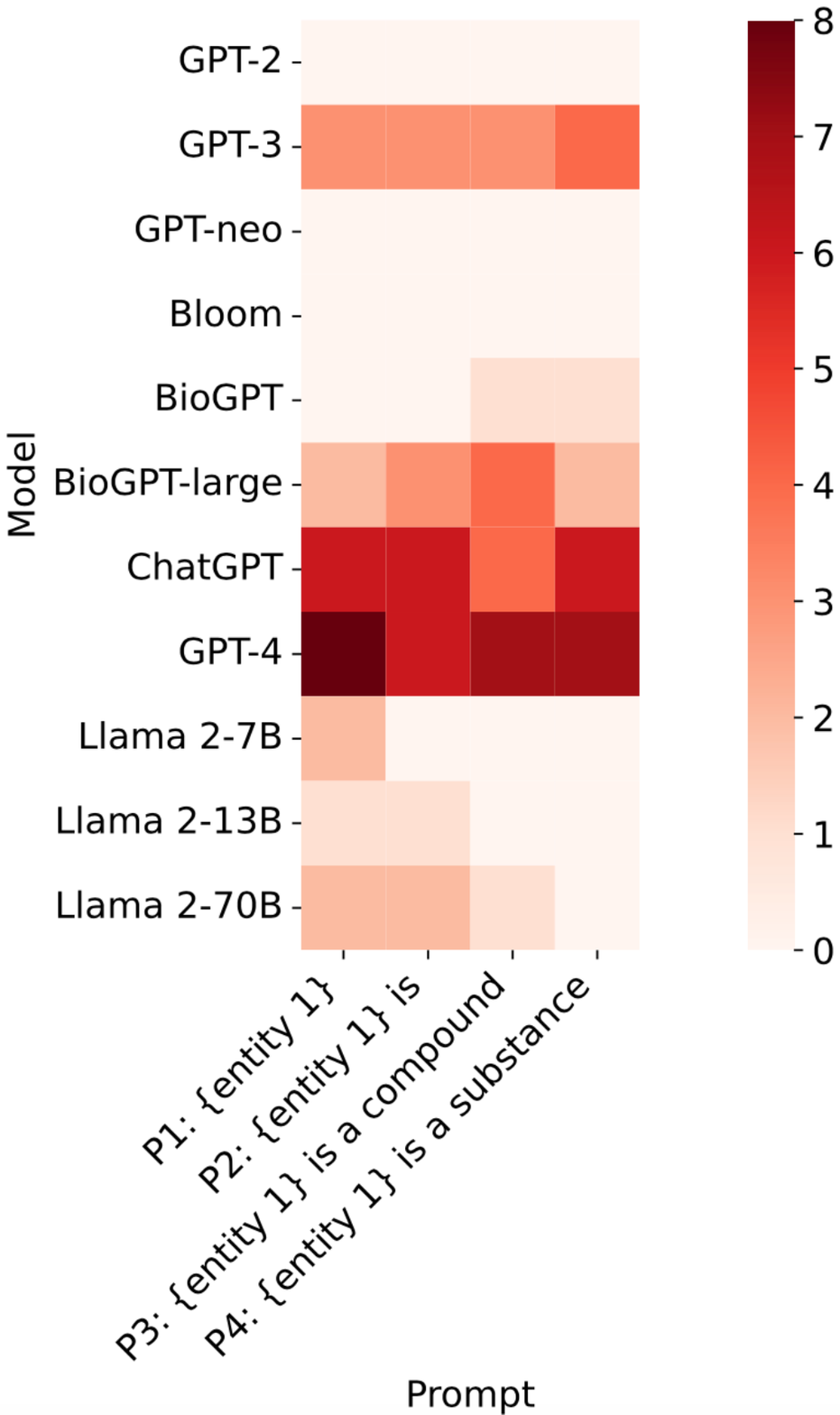}  
        \label{fig:sub2}
    \end{subfigure}
    \caption{Heatmap showing the factuality (STEP 2) (A) and specificity (STEP 3) (B) of outputs across different models and prompts in Task 1. The x-axis represents the prompts P1-P4, while the y-axis lists the AI models used in the study. The intensity of the color indicates the effectiveness of a given prompt, with darker shades representing responses with higher factuality (A) or factual responses with higher specificity (B).}
    \label{fig:heatmap1}
\end{figure}

\begin{figure}[htbp]
    \centering
    \begin{subfigure}[t]{0.45\textwidth}  
        \centering
        \subcaption*{A)}
        \includegraphics[width=\linewidth]{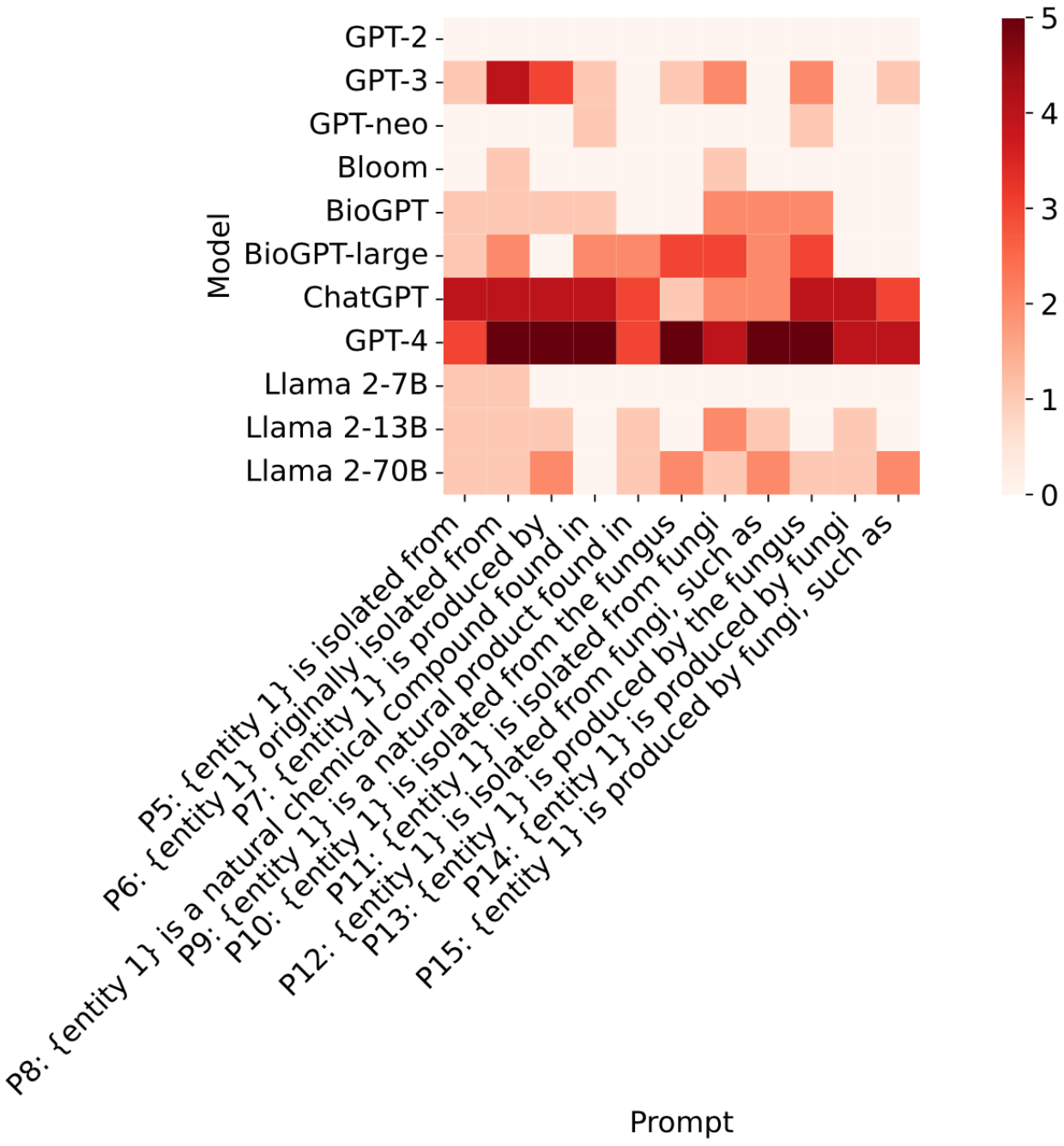}  
        \label{fig:sub1}
    \end{subfigure}
    \hspace{0.05\textwidth}  
    \begin{subfigure}[t]{0.45\textwidth}  
        \centering
        \subcaption*{B)}
        \includegraphics[width=\linewidth]{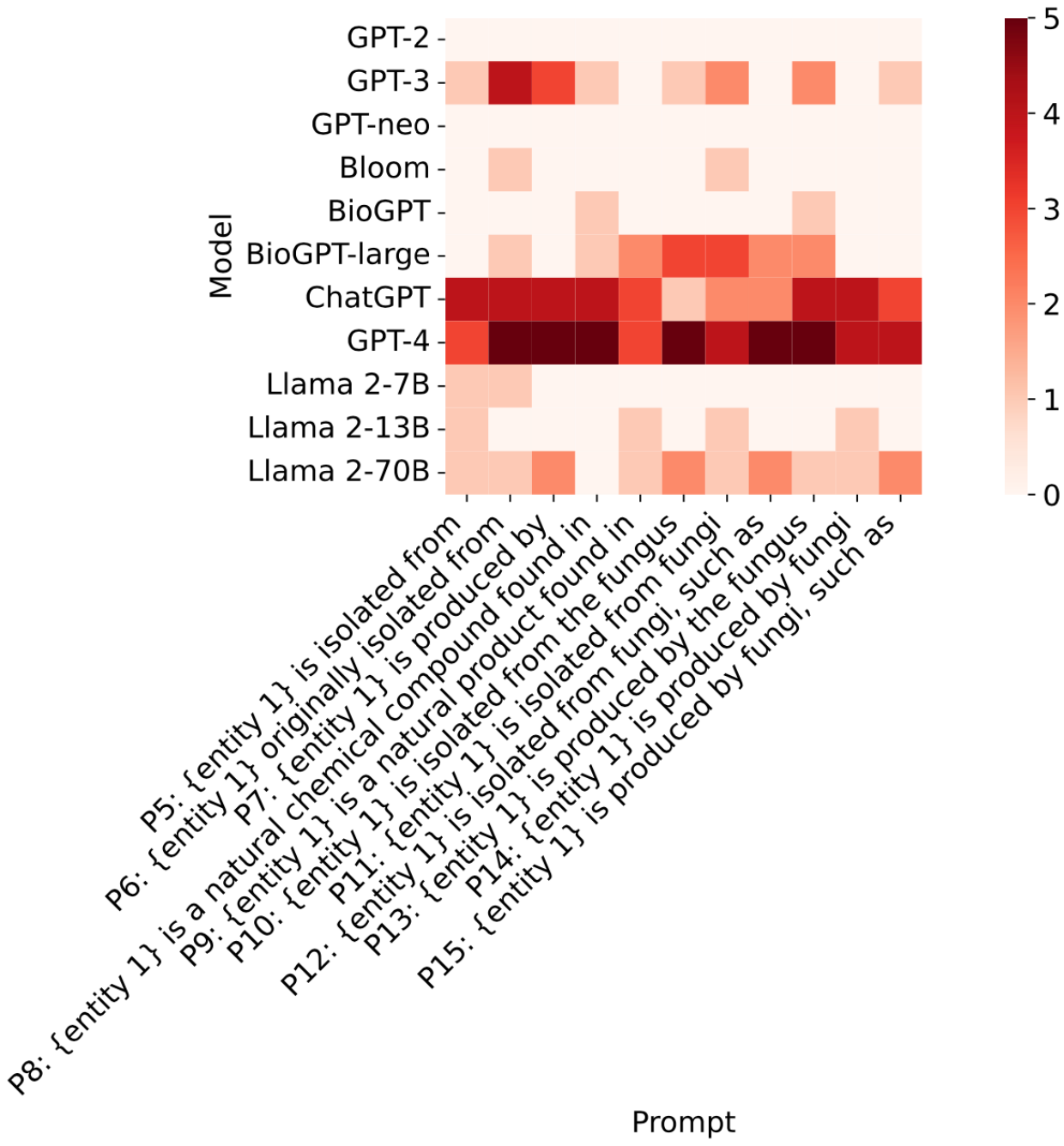}  
        \label{fig:sub2}
    \end{subfigure}
    \caption{Heatmap showing the factuality (STEP 2B) (A) and specificity (STEP 3) (B) of outputs across different models and prompts in Task 2. The x-axis represents the prompts P5-P15, while the y-axis lists the AI models used in the study. The intensity of the color indicates the effectiveness of a given prompt, with darker shades representing responses with higher factuality (A) or factual responses with higher specificity (B).}
    \label{fig:heatmap2}
\end{figure}

\newcounter{app_table}
\setcounter{app_table}{1}

\captionsetup[table]{labelformat=empty}

\renewcommand{\thetable}{A.\arabic{app_table}}

\setcounter{table}{1}

\begin{table}[h]
\caption[tab]{Table \thetable. The dataset for the experiments contains ten chemical compounds (chem name): five with (antibiotic activity = 1) and five without antibiotic activity (antibiotic activity = 0). For each relation of a selected chemical compound with a fungus (fungi name) there is a number of references (nb ref) describing the relation based on PubChem.}
\refstepcounter{app_table} 
\label{tab:short_dataset}
\resizebox{\columnwidth}{!}{%
\begin{tabular}{@{}llllllll@{}}
\toprule
\textbf{fungi id} & \textbf{\begin{tabular}[c]{@{}l@{}}fungi name \\ \textit{entity 2}\end{tabular}} & \textbf{family name}          & \textbf{pubchem id} & \textbf{\begin{tabular}[c]{@{}l@{}}chem name \\ \textit{entity 1}\end{tabular}} & \textbf{\begin{tabular}[c]{@{}l@{}}antibiotic \\ activity \\ 0/1\end{tabular}} & \textbf{nb ref} & \textbf{\begin{tabular}[c]{@{}l@{}}Publication year of \\ the latest reference\end{tabular}} \\ \midrule
211776            & Aspergillus fumigatus                                                   & \textit{Aspergillaceae}       & 403923              & Fumitremorgin C                                                        & 0                                                                              & 10              & 2015                                                                                         \\
119834            & Alternaria alternata                                                    & \textit{Pleosporaceae}        & 5359485             & Alternariol                                                            & 0                                                                              & 9               & 2012                                                                                         \\
148413            & Ganoderma lucidum                                                       & \textit{Polyporaceae}         & 444679              & Ergosterol                                                             & 0                                                                              & 6               & 2009                                                                                         \\
100745            & Ganoderma australe                                                      & \textit{Polyporaceae}         & 444679              & Ergosterol                                                             & 0                                                                              & 3               & 2004                                                                                         \\
119872            & Ganoderma applanatum                                                    & \textit{Polyporaceae}         & 444679              & Ergosterol                                                             & 0                                                                              & 3               & 2014                                                                                         \\
504340            & Ophiocordyceps sinensis                                                 & \textit{Ophiocordycipitaceae} & 444679              & Ergosterol                                                             & 0                                                                              & 3               & 2018                                                                                         \\
182069            & Aspergillus nidulans                                                    & \textit{Aspergillaceae}       & 444679              & Ergosterol                                                             & 0                                                                              & 3               & 2009                                                                                         \\
298052            & Gymnopilus spectabilis                                                  & \textit{Strophariaceae}       & 444679              & Ergosterol                                                             & 0                                                                              & 2               & 2005                                                                                         \\
250886            & Phaeolepiota aurea                                                      & \textit{Agaricaceae}          & 444679              & Ergosterol                                                             & 0                                                                              & 2               & 1991                                                                                         \\
315778            & Hypsizygus marmoreus                                                    & \textit{Agaricaceae}          & 444679              & Ergosterol                                                             & 0                                                                              & 2               & 2006                                                                                         \\
236989            & Monilinia fructicola                                                    & \textit{Sclerotiniaceae}      & 121225493           & (+)-Chloromonilicin                                                    & 0                                                                              & 2               & 2011                                                                                         \\
247956            & Penicillium aurantiogriseum                                             & \textit{Aspergillaceae}       & 10892066            & Conidiogenone                                                          & 0                                                                              & 2               & 2003                                                                                         \\ \midrule
257047            & Cephalosporium aphidicola                                               & \textit{Cordycipitaceae}      & 457964              & Aphidicolin                                                            & 1                                                                              & 10              & 2004                                                                                         \\
303853            & Pleospora betae                                                         & \textit{Pleosporaceae}        & 457964              & Aphidicolin                                                            & 1                                                                              & 5               & 2018                                                                                         \\
337192            & Pleospora bjoerlingii                                                   & \textit{Neocamarosporiaceae}  & 457964              & Aphidicolin                                                            & 1                                                                              & 5               & 2016                                                                                         \\
284309            & Aspergillus niger                                                       & \textit{Aspergillaceae}       & 5748546             & Flavasperone                                                           & 1                                                                              & 6               & 2004                                                                                         \\
815927            & Albifimbria verrucaria                                                  & \textit{Stachybotryaceae}     & 6326658             & Verrucarin A                                                           & 1                                                                              & 5               & 2012                                                                                         \\
815989            & Paramyrothecium roridum                                                 & \textit{Stachybotryaceae}     & 6326658             & Verrucarin A                                                           & 1                                                                              & 3               & 2019                                                                                         \\
245274            & Myrothecium verrucaria                                                  & \textit{Stachybotryaceae}     & 6326658             & Verrucarin A                                                           & 1                                                                              & 2               & 2012                                                                                         \\
283055            & Dendrodochium toxicum                                                   & \textit{Bionectriaceae}       & 6326658             & Verrucarin A                                                           & 1                                                                              & 2               & 2004                                                                                         \\
360246            & Tapinella atrotomentosa                                                 & \textit{Tapinellaceae}        & 99148               & Atromentin                                                             & 1                                                                              & 2               & 2007                                                                                         \\
340026            & Thelephora aurantiotincta                                               & \textit{Thelephoraceae}       & 99148               & Atromentin                                                             & 1                                                                              & 2               & 2005                                                                                         \\
504278            & Ophiocordyceps heteropoda                                               & \textit{Ophiocordycipitaceae} & 6438394             & Myriocin                                                               & 1                                                                              & 2               & 2009                                                                                         \\ \bottomrule
\end{tabular}%
}
\end{table}

\begin{landscape}
\begin{table}[]
\caption{Table \thetable. Summary of the evaluated Large Language Models ordered by the release date.}
\refstepcounter{app_table} 
\label{tab:generative_models_summary}
\centering
\resizebox{0.9\columnwidth}{!}{%
\begin{tabular}{@{}cccccccccc@{}}
\toprule
{\color[HTML]{000000} \textbf{Model}} & {\color[HTML]{000000} \textbf{Lab}} & {\color[HTML]{000000} \textbf{Training Corpus}}                                                                                                                                                                                                              & {\color[HTML]{000000} \textbf{Vocabulary size}} & {\color[HTML]{000000} \textbf{\begin{tabular}[c]{@{}c@{}}Model size\\ (parameters)\\used/available\end{tabular}}}                     & {\color[HTML]{000000} \textbf{\begin{tabular}[c]{@{}c@{}}Layers (L),\\ hidden size (H),\\heads (A)\end{tabular}}} & {\color[HTML]{000000} \textbf{\begin{tabular}[c]{@{}c@{}}Maximum input\\ sequence\\length (tokens) \end{tabular}}} & {\color[HTML]{000000} \textbf{\begin{tabular}[c]{@{}c@{}}Date of the\\ model release*\end{tabular}}} & {\color[HTML]{000000} \textbf{\begin{tabular}[c]{@{}c@{}}Reference\end{tabular}}}                 \\ \midrule
{\color[HTML]{000000} GPT-2}          & {\color[HTML]{000000} OpenAI}       & {\color[HTML]{000000} \begin{tabular}[c]{@{}c@{}}WebText: 40 GB of text, 8M documents, \\ from 45M webpages upvoted on Reddit \\ (all Wikipedia pages removed)\end{tabular}}                                                       & {\color[HTML]{000000} 50,257}                   & {\color[HTML]{000000} \begin{tabular}[c]{@{}c@{}}124M/1.5B \end{tabular}} & {\color[HTML]{000000} L=24, H=1024, A=16} & {\color[HTML]{000000} 1024} & {\color[HTML]{000000} Feb 2019}      & {\color[HTML]{000000} \cite{radford2019language}}                                                                                                                            \\
{\color[HTML]{000000} GPT-3}          & {\color[HTML]{000000} OpenAI}       & {\color[HTML]{000000} \begin{tabular}[c]{@{}c@{}}570 GB plaintext, 0.4 trillion tokens. \\ (410B CommonCrawl, 19B WebText2, \\ 3B English Wikipedia, and two books \\ corpora (12B Books1 and 55B Books2))\end{tabular}}                                     & {\color[HTML]{000000} }               & {\color[HTML]{000000} 175B}                                                                                                                  & {\color[HTML]{000000} L=96, H=12288, A=96} & {\color[HTML]{000000} 2048} & {\color[HTML]{000000} June 2020}     & {\color[HTML]{000000} \cite{brown2020language} }                                                                                                                                    \\
{\color[HTML]{000000} GPT-neo}        & {\color[HTML]{000000} EleutherAI}   & {\color[HTML]{000000} \begin{tabular}[c]{@{}c@{}}The Pile - a large scale curated \\ dataset created by EleutherAI  \\ 825GB, 380 billion tokens\end{tabular}} & {\color[HTML]{000000} 50,257}                   & {\color[HTML]{000000} 1.3B/2.7B}                                                                                                                 & {\color[HTML]{000000} L=24, H=2048, A=16} & {\color[HTML]{000000} 2048}  & {\color[HTML]{000000} March 2021}    & {\color[HTML]{000000} \cite{gao2020pile} }                                                                     \\
{\color[HTML]{000000} Bloom}          & {\color[HTML]{000000} BigScience}   & {\color[HTML]{000000} \begin{tabular}[c]{@{}c@{}}45 natural languages\\ 12 programming languages\\ In 1.5TB of pre-processed text, \\ converted into 350B unique tokens\end{tabular}}                                                                     & {\color[HTML]{000000} 250,680}                  & {\color[HTML]{000000} 560M-3B/176B}                                                                                                                & {\color[HTML]{000000} L=24, H=1024, A=16} & {\color[HTML]{000000} 2048}   & {\color[HTML]{000000} July 2022}     & {\color[HTML]{000000} \cite{BLOOMmodelcard} }                                                                        \\
{\color[HTML]{000000} BioGPT}         & {\color[HTML]{000000} Microsoft}    & {\color[HTML]{000000} 15M PubMed abstracts (trained from scratch)}                                                                                                                                                                                                     & {\color[HTML]{000000} 42,384}                   & {\color[HTML]{000000} 347M/347M}                                                                                                        & {\color[HTML]{000000} L=24, H=1024, A=16} & {\color[HTML]{000000} 1024}     & {\color[HTML]{000000} Oct 2022}      & {\color[HTML]{000000} \cite{LuoBioGPT}}                                                                         \\
{\color[HTML]{000000} BioGPT-large}   & {\color[HTML]{000000} Microsoft}    & {\color[HTML]{000000} 15M PubMed abstracts (trained from scratch)}                                                                                                                                                                                                     & {\color[HTML]{000000} 42,384}                   & {\color[HTML]{000000} 1.5B/1.5B}                                                                                                           & {\color[HTML]{000000} L=24, H=1024, A=16} & {\color[HTML]{000000} 2048}        & {\color[HTML]{000000} Oct 2022}      & {\color[HTML]{000000} \cite{LuoBioGPT}}                                                                     \\
{\color[HTML]{000000} ChatGPT}        & {\color[HTML]{000000} OpenAI}       & {\color[HTML]{000000} unavailable}                                                                                                                                                                                                                                      & {\color[HTML]{000000} }                         & {\color[HTML]{000000} 20B (175B)}                                                                                                          & {\color[HTML]{000000} } & {\color[HTML]{000000} 4096}   & {\color[HTML]{000000} Nov 2022}      & {\color[HTML]{000000} }                                                                                                                           \\
{\color[HTML]{000000} GPT-4}        & {\color[HTML]{000000} OpenAI}       & {\color[HTML]{000000} unavailable}                                                                                                                                                                                                                                      & {\color[HTML]{000000} }                         & {\color[HTML]{000000} }                                                                                                            & {\color[HTML]{000000}L=48, H=12288, A=96 } & {\color[HTML]{000000} 8192}  & {\color[HTML]{000000} Feb 2023}      & {\color[HTML]{000000} }                                                                                                                                 \\
{\color[HTML]{000000} }        & {\color[HTML]{000000} }       & {\color[HTML]{000000} }                                                                                                                                                                                                                                      & {\color[HTML]{000000} }                         & {\color[HTML]{000000} 7B}                                                                                                         & {\color[HTML]{000000} L=32, H=4096, A=32} & {\color[HTML]{000000} }      & {\color[HTML]{000000} }      & {\color[HTML]{000000} }                                                                                                                                 \\
{\color[HTML]{000000} Llama 2}        & {\color[HTML]{000000} Meta AI}       & {\color[HTML]{000000}  unavailable}                                                                                                                                                                                                                                      & {\color[HTML]{000000} }                         & {\color[HTML]{000000} 13B}                                                                                                        & {\color[HTML]{000000} L=40, H=5120, A=40} & {\color[HTML]{000000} 4096}       & {\color[HTML]{000000} July 2023}      & {\color[HTML]{000000} \cite{touvron2023llama}}                                                                                                                                 \\
{\color[HTML]{000000} }        & {\color[HTML]{000000} }       & {\color[HTML]{000000} }                                                                                                                                                                                                                                      & {\color[HTML]{000000} }                         & {\color[HTML]{000000} 70B}                                                                                                          & {\color[HTML]{000000} L=80, H=8192, A=64} & {\color[HTML]{000000} }     & {\color[HTML]{000000} }      & {\color[HTML]{000000} }                                                                                                                                 \\\bottomrule
\end{tabular}%
}
\begin{tablenotes}\footnotesize
\item[*] *The ‘Date of the model release’ is based on its first arXiv version (if exists) or official public release.
\end{tablenotes}
\end{table}
\end{landscape}

\begin{table}[]
\caption{Table \thetable. Examples of generated text for selected models, selected compounds as input, and selected prompts. In the 'Comment' column there are correct answers or comments by the authors referring to semantic coherence or factuality of generated text.}
\refstepcounter{app_table} 
\label{tab:text_generation_short}
\resizebox{\columnwidth}{!}{%
\begin{tabular}{@{}clllll@{}}
\toprule
\multicolumn{1}{l}{Model}        & Input               & Prompt                                                               & Generated Text                                                                                                                                                                                                                                                                                                                                                                                                                                                                                                                                                                                                                                                                                                                                                                                                                                                                                                                                                                                                                                                                                                                                                                                                                                   & Comment                                                                                                                                                                                                                                                            & \begin{tabular}[c]{@{}l@{}}STEP 2 \\ Factuality\end{tabular}  \\ \midrule
\multirow{13}{*}{BioGPT}         & Aphidicolin         & \{entity 1\} is                                                        & \begin{tabular}[c]{@{}l@{}}Aphidicolin is  a potent inhibitor of DNA polymerase \\ alpha and beta, but not of DNA polymerase gamma.\end{tabular}                                                                                                                                                                                                                                                                                                                                                                                                                                                                                                                                                                                                                                                                                                                                                                                                                                                                                                                                                                                                                                                                                                 & \begin{tabular}[c]{@{}l@{}}Aphidicolin is a potent inhibitor of \\ both host cell DNA polymerase alpha \\ and herpes simplex virus (HSV)-induced \\ DNA polymerase but has no effect on \\ DNA polymerases beta and gamma \\ of host cells. {[}ref{]}\end{tabular}                                                                & 0                                                            \\
                                 & Ergosterol          & \{entity 1\} is                                                        & \begin{tabular}[c]{@{}l@{}}Ergosterol is  a natural product that has been shown \\ to have a wide range of biological activities.\end{tabular}                                                                                                                                                                                                                                                                                                                                                                                                                                                                                                                                                                                                                                                                                                                                                                                                                                                                                                                                                                                                                                                                                                   & \begin{tabular}[c]{@{}l@{}}General description which was repeated \\ by the model several times.\end{tabular}                                                                                                                                                      & 1                                                             \\
                                 & Aphidicolin         & \begin{tabular}[c]{@{}l@{}}\{entity 1\} is \\ a compound\end{tabular}  & \begin{tabular}[c]{@{}l@{}}Aphidicolin is a compound  that inhibits DNA \\ replication by inhibiting the enzyme DNA \\ polymerase alpha.\end{tabular}                                                                                                                                                                                                                                                                                                                                                                                                                                                                                                                                                                                                                                                                                                                                                                                                                                                                                                                                                                                                                                                                                   &                                                                                                                                                                                                                                                                    & 1                                                               \\
                                 & Aphidicolin         & \begin{tabular}[c]{@{}l@{}}\{entity 1\} is \\ a substance\end{tabular} & \begin{tabular}[c]{@{}l@{}}Aphidicolin is a substance  that inhibits DNA \\ replication by blocking the replicative DNA \\ polymerase.\end{tabular}                                                                                                                                                                                                                                                                                                                                                                                                                                                                                                                                                                                                                                                                                                                                                                                                                                                                                                                                                                                                                                                                                     &                                                                                                                                                                                                                                                                    & 1                                                               \\
                                 & Flavasperone        & \begin{tabular}[c]{@{}l@{}}\{entity 1\} is \\ a compound\end{tabular}  & \begin{tabular}[c]{@{}l@{}}Flavasperone is a compound  with a wide range \\ of pharmacological properties.\end{tabular}                                                                                                                                                                                                                                                                                                                                                                                                                                                                                                                                                                                                                                                                                                                                                                                                                                                                                                                                                                                                                                                                                                                          & \begin{tabular}[c]{@{}l@{}}General description which was \\ repeated by the model several times.\end{tabular}                                                                                                                                                      & 1                                                      \\
                                 & Verrucarin A        & \begin{tabular}[c]{@{}l@{}}\{entity 1\} is \\ a compound\end{tabular}  & \begin{tabular}[c]{@{}l@{}}Verrucarin A is a compound  with a unique structure \\ and a unique mechanism of action.\end{tabular}                                                                                                                                                                                                                                                                                                                                                                                                                                                                                                                                                                                                                                                                                                                                                                                                                                                                                                                                                                                                                                                                                                                 & \begin{tabular}[c]{@{}l@{}}General description which was \\ repeated by the model several times.\end{tabular}                                                                                                                                                      & 0                                                            \\
                                 & Fumitremorgin C     & \begin{tabular}[c]{@{}l@{}}\{entity 1\} is \\ a compound\end{tabular}  & \begin{tabular}[c]{@{}l@{}}Fumitremorgin C is a compound  with a unique \\ structure and a unique mode of action.\end{tabular}                                                                                                                                                                                                                                                                                                                                                                                                                                                                                                                                                                                                                                                                                                                                                                                                                                                                                                                                                                                                                                                                                                                   & \begin{tabular}[c]{@{}l@{}}General description which was \\ repeated by the model several times.\end{tabular}                                                                                                                                                      & 0                                                            \\
                                 & Alternariol         & \begin{tabular}[c]{@{}l@{}}\{entity 1\} is \\ a compound\end{tabular}  & \begin{tabular}[c]{@{}l@{}}Alternariol is a compound  with a wide range \\ of biological activities.\end{tabular}                                                                                                                                                                                                                                                                                                                                                                                                                                                                                                                                                                                                                                                                                                                                                                                                                                                                                                                                                                                                                                                                                                                                & \begin{tabular}[c]{@{}l@{}}General description which was \\ repeated by the model several times.\end{tabular}                                                                                                                                                      & 1                                                      \\
                                 & Fumitremorgin C     & \begin{tabular}[c]{@{}l@{}}\{entity 1\} is \\ a substance\end{tabular} & \begin{tabular}[c]{@{}l@{}}Fumitremorgin C is a substance  that is produced \\ by the fungus Aspergillus terreus.\end{tabular}                                                                                                                                                                                                                                                                                                                                                                                                                                                                                                                                                                                                                                                                                                                                                                                                                                                                                                                                                                                                                                                                                                                   & \begin{tabular}[c]{@{}l@{}}Fumitremorgin C is not produced \\ by the fungus Aspergillus terreus.\end{tabular}                                                                                                                                                      & 0                                                            \\
                                 & Ergosterol          & \begin{tabular}[c]{@{}l@{}}\{entity 1\} is \\ a substance\end{tabular} & \begin{tabular}[c]{@{}l@{}}Ergosterol is a substance  that is produced by \\ the fungus Ergoderma lucidum.\end{tabular}                                                                                                                                                                                                                                                                                                                                                                                                                                                                                                                                                                                                                                                                                                                                                                                                                                                                                                                                                                                                                                                                                                                          & \begin{tabular}[c]{@{}l@{}}Model generates new name of fungus. \\ There is no Ganoderma lucidum \\ in the literature. {[}ref{]}\end{tabular}                                                                                                                       & 0                                                            \\
                                 & Conidiogenone       & \begin{tabular}[c]{@{}l@{}}\{entity 1\} is \\ a substance\end{tabular} & \begin{tabular}[c]{@{}l@{}}Conidiogenone is a substance  isolated from \\ the fungus Aspergillus versicolor.\end{tabular}                                                                                                                                                                                                                                                                                                                                                                                                                                                                                                                                                                                                                                                                                                                                                                                                                                                                                                                                                                                                                                                                                                                        & \begin{tabular}[c]{@{}l@{}}Conidiogenone is not isolated from \\ the fungus Aspergillus versicolor.\end{tabular}                                                                                                                                                   & 0                                                            \\
                                 & (+)-Chloromonilicin & \begin{tabular}[c]{@{}l@{}}\{entity 1\} is \\ a substance\end{tabular} & \begin{tabular}[c]{@{}l@{}}(+)-Chloromonilicin is a substance that inhibits the \\ growth of the fungus Fusarium oxysporum f.\end{tabular}                                                                                                                                                                                                                                                                                                                                                                                                                                                                                                                                                                                                                                                                                                                                                                                                                                                                                                                                                                                                                                                                                                       & \begin{tabular}[c]{@{}l@{}}(+)-Chloromonilicin doesn't inhibit the \\ growth of the fungus Fusarium \\ oxysporum f.\end{tabular}                                                                                                                                   & 0                                                            \\
                                 & Verrucarin A        & \begin{tabular}[c]{@{}l@{}}\{entity 1\} is \\ a substance\end{tabular} & \begin{tabular}[c]{@{}l@{}}Verrucarin A is a substance  that is produced by \\ the fungus Aspergillus versicolor.\end{tabular}                                                                                                                                                                                                                                                                                                                                                                                                                                                                                                                                                                                                                                                                                                                                                                                                                                                                                                                                                                                                                                                                                                                   & \begin{tabular}[c]{@{}l@{}}Verrucarin A is not produced by the \\ fungus Aspergillus versicolor.\end{tabular}                                                                                                                                                      & 0                                                            \\ \midrule
\multicolumn{1}{l}{BioGPT-Large} & (+)-Chloromonilicin & \{entity 1\} is                                                        & \begin{tabular}[c]{@{}l@{}}(+)-Chloromonilicin is  a new antibiotic produced \\ by a strain of Streptomyces. It is active against \\ Gram-positive bacteria and fungi. The structure \\ of (+) -chloromon\end{tabular}                                                                                                                                                                                                                                                                                                                                                                                                                                                                                                                                                                                                                                                                                                                                                                                                                                                                                                                                                                                                                           & \begin{tabular}[c]{@{}l@{}}Chloromonilicin is a natural \\ product found in Alternaria \\ sonchi and Monilinia fructicola .\end{tabular}                                                                                                                           & 0                                                            \\ \midrule
\multirow{4}{*}{ChatGPT}         & Aphidicolin         & \{entity 1\}                                                           & \begin{tabular}[c]{@{}l@{}}Aphidicolin is a naturally occurring compound that \\ was originally isolated from the fungus Cephalosporium \\ aphidicola. It is a potent inhibitor of DNA synthesis and \\ is commonly used as a research tool to study DNA replication.\\ \\ Aphidicolin inhibits the activity of DNA polymerases, \\ which are the enzymes responsible for synthesizing \\ new DNA strands during DNA replication. By blocking \\ this process, aphidicolin prevents cells from dividing and \\ proliferating. This makes it useful for studying the effects \\ of DNA replication inhibition on cellular processes such \\ as DNA repair, cell cycle progression, and gene expression.\\ \\ Aphidicolin has also been investigated as a potential \\ anti-cancer agent, as many cancer cells rely on rapid \\ DNA replication to support their growth and survival. \\ However, its clinical use has been limited by its toxicity \\ and side effects.\\ \\ Overall, aphidicolin is a valuable research tool for studying \\ DNA replication and has potential applications in cancer \\ treatment, but further research is needed to better understand \\ its mechanisms of action and potential therapeutic uses.\end{tabular} &                                                                                                                                                                                                                                                                    & 1 \\
                                 & Aphidicolin         & \{entity 1\} is                                                        & \begin{tabular}[c]{@{}l@{}}Aphidicolin is a naturally occurring compound that is \\ commonly used as a research tool to study DNA replication. \\ It is a potent inhibitor of DNA synthesis and works by \\ blocking the activity of DNA polymerases, which are \\ the enzymes responsible for synthesizing new DNA strands \\ during DNA replication. By inhibiting DNA replication, \\ aphidicolin prevents cells from dividing and proliferating, \\ making it useful for studying the effects of DNA replication \\ inhibition on cellular processes such as DNA repair, cell cycle \\ progression, and gene expression. Aphidicolin has also been \\ investigated as a potential anti-cancer agent, but its clinical \\ use has been limited by its toxicity and side effects.\end{tabular}                                                                                                                                                                                                                                                                                                                                                                                                                                                 &                                                                                                                                                                                                                                                                    & 1                                                            \\
                                 & Aphidicolin         & \begin{tabular}[c]{@{}l@{}}\{entity 1\} is \\ a compound\end{tabular}  & \begin{tabular}[c]{@{}l@{}}Yes, aphidicolin is a compound. It is a naturally\\  occurring tetracyclic diterpene, which means it is \\ a type of organic molecule made up of four fused \\ rings and derived from the isoprene unit. The chemical \\ structure of aphidicolin consists of 18 carbon atoms, \\ 28 hydrogen atoms, and 2 oxygen atoms.\end{tabular}                                                                                                                                                                                                                                                                                                                                                                                                                                                                                                                                                                                                                                                                                                                                                                                                                                                                                 & The chemical formula was incorrect                                                                                                                                                                                                                                 & 0                                                            \\
                                 & Aphidicolin         & \begin{tabular}[c]{@{}l@{}}\{entity 1\} is \\ a substance\end{tabular} & \begin{tabular}[c]{@{}l@{}}Yes, aphidicolin is a substance. It is a naturally occurring \\ organic compound that is found in certain fungi, including \\ Cephalosporium aphidicola. It can also be synthesized \\ chemically in the laboratory. As a substance, aphidicolin \\ is used as a research tool to study DNA replication and \\ as a potential anti-cancer agent, although its clinical use \\ has been limited due to its toxicity and side effects.\end{tabular}                                                                                                                                                                                                                                                                                                                                                                                                                                                                                                                                                                                                                                                                                                                                                                     &                                                                                                                                                                                                                                                                    & 1                                                            \\ \bottomrule
\end{tabular}%
}
\end{table}

\newpage

\begin{table}[]
\caption{Table \thetable. Examples of generated text for selected input: chemical compound  \textit{Myriocin} and a simple prompt: '\{{\color{blue}entity 1}\}' for all eleven models analyzed in the paper. Factuality of generated text are provided for all generated responses.}
\refstepcounter{app_table} 
\label{tab:text_generation_myriocin}
\resizebox{1.0\columnwidth}{!}{%
\begin{tabular}{@{}llll@{}}
\toprule
Model        & Generated Text                                                                                                                                                                                                                                                                                                                                                                                                                                                                                                                                                                                                                                                                                                                                                                                                                                                                                                                                                                                                                                                                                                                                                                                                                                                                       & \begin{tabular}[c]{@{}l@{}}STEP 2 \\ Factuality \end{tabular} \\ \midrule
GPT-2        & \begin{tabular}[c]{@{}l@{}}Myriocin ¬†(Kaihuil) In this series, we see the relationship between three main protagonists, Kaihua, Kihua and I \\ \\ \end{tabular}                                                                                                                                                                                                                                                                                                                                                                                                                                                                                                                                                                                                                                                                                                                                                                                                                                                                                                                                                                                                                                                                                                                               & 0                                                            \\
GPT-3        & \begin{tabular}[c]{@{}l@{}}Myriocin is an antifungal drug used to treat fungal infections. It is derived from the fungus Streptomyces myriocin, \\and is used to treat a variety of fungal infections, including Candida\\ \\ \end{tabular}                                                                                                                                                                                                                                                                                                                                                                                                                                                                                                                                                                                                                                                                                                                                                                                                                                                                                                                                                                                                                                                    & 0  \\
gpt-neo      & Myriocin                                                                                                                                                                                                                                                                                                                                                                                                                                                                                                                                                                                                                                                                                                                                                                                                                                                                                                                                                                                                                                                                                                                                                                                                                                                                                     & 0                                                            \\ \\
bloom        & \begin{tabular}[c]{@{}l@{}}Myriocin ÊäóËÇøÁò§ËçØ ÊäóËÇøÁò§ËçØ ÊäóËÇøÁò§ËçØ  ÊäóËÇøÁò§ËçØ ÊäóËÇøÁò§ËçØ ÊäóËÇøÁò§ËçØ \\ÊäóËÇøÁò§ËçØ ÊäóËÇøÁò§ \\ \\ \end{tabular}                                                                                                                                                                                                                                                                                                                                                                                                                                                                                                                                                                                                                                                                                                                                                                                                                                                                                                                                                                                                                                                                                                                                  & 0                                                            \\
BioGPT       & \begin{tabular}[c]{@{}l@{}}Myriocin , a novel peptide from the venom of the  scorpion Leiurus quinquestriatus hebraeus.\\ \\ \end{tabular}                                                                                                                                                                                                                                                                                                                                                                                                                                                                                                                                                                                                                                                                                                                                                                                                                                                                                                                                                                                                                                                                                                                                                      & 0                                                            \\
BioGPT-large & \begin{tabular}[c]{@{}l@{}}Myriocin , a specific inhibitor of serine palmitoyltransferase, was used to inhibit de novo sphingolipid \\synthesis. The results showed that myriocin treatment significantly reduced the levels of \\ \\ \end{tabular}                                                                                                                                                                                                                                                                                                                                                                                                                                                                                                                                                                                                                                                                                                                                                                                                                                                                                                                                                                                                                                    & 1                                                            \\
ChatGPT      & \begin{tabular}[c]{@{}l@{}}Myriocin is a natural product and a type of mycotoxin that is produced by several species of fungi, including \\Isaria sinclairii and Myriococcum albomyces. It was first isolated from the fungus Myriococcum albomyces in \\1994. Myriocin is also known as thermozymocidin, ISP-1 or Penicillium mycotoxin, and it has been shown to have \\a range of biological activities. Myriocin is a potent inhibitor of serine palmitoyltransferase, which is an enzyme \\involved in the synthesis of sphingolipids. Sphingolipids are important structural components of cell membranes \\and play a role in signaling pathways. By inhibiting serine palmitoyltransferase, Myriocin disrupts sphingolipid \\synthesis, which can lead to apoptosis (programmed cell death) in certain cells.\\ Myriocin has been studied for its potential use as a therapeutic agent for a variety of conditions, including cancer, \\inflammation, and autoimmune diseases. It has also been investigated for its potential as an immunosuppressive \\agent in the treatment of organ transplantation and autoimmune diseases. Additionally, Myriocin has been found \\to exhibit antifungal and antiparasitic activity.\\ \\ \end{tabular}          & 1                                                    \\
GPT-4      & \begin{tabular}[c]{@{}l@{}}Myriocin, also known as ISP-1 or thermozymocidin, is a naturally occurring compound with potent \\immunosuppressive and antifungal properties. It is derived from the fungus Mycelia sterilia, specifically the Isaria \\sinclairii species. Myriocin is a structural analog of sphingosine, a component of sphingolipids, which are essential \\ components of cell membranes in eukaryotes.\\ Myriocin acts by inhibiting serine palmitoyltransferase (SPT), a key enzyme in the de novo biosynthesis of \\sphingolipids. This inhibition disrupts the synthesis of sphingolipids, leading to alterations in cell membrane \\structure and function. Consequently, myriocin has the potential to suppress various immune responses and limit \\ the growth of certain fungal pathogens.\\ \\ \end{tabular}          & 1                                                             \\
Llama 2-7B   & \begin{tabular}[c]{@{}l@{}}is a new antibiotic that has been developed to treat bacterial infections. It is effective against many types of bacteria, \\including those that are resistant to other antibiotics. Myriocin works by interfering with the bacteria‚\\ Äôs ability to make proteins. This prevents the bacteria from growing and multiplying, which eventually kills them. \\Myriocin is available in both oral and intravenous forms.\\ \\ \end{tabular} & 0                \\
Llama 2-13B  & \begin{tabular}[c]{@{}l@{}}is a natural product isolated from the marine sponge Mycale hentscheli. It has been shown to have antitumor \\activity in vitro and in vivo, but its mechanism of action remains unknown. In this study, we investigated \\the effect of myriocin on cell cycle progression and apoptosis in human breast cancer cells.\\ \\ \end{tabular}  & 0                 \\
Llama 2-70B  & \begin{tabular}[c]{@{}l@{}}(CAS 10238-79-6) World Market Report 2019 aims at providing comprehensive data on myriocin market globally \\and regionally (Europe, Asia, North America, Latin America etc.). The report includes myriocin description, \\covers its application areas, manufacturing methods, patterns. It captures myriocin market trends, pays close \\attention to myriocin manufacturers and names suppliers. Furthermore, myriocin prices in regional markets \\can be found in the report with regards to countries and companies. In addition to the above the report determines \\ myriocin consumers. 3. MYRIOCIN MARKET WORLDWIDE. MANUFACTURERS AND SUPPLIERS OF \\MYRIOCIN (INCLUDING CONTACT DETAILS).\end{tabular} & 0                 \\ \bottomrule

\end{tabular}%
}
\end{table}

\newpage

\begin{landscape}
\begin{table}[]
\caption{Table \thetable. Generated fungus name in Task 2: \textit{chemical-fungus relation determination via entity generation task} for one chemical compound \textit{Alternariol} for 11 prompts (P5-P15). Fungus names that are generated correctly in bold.}
\refstepcounter{app_table} 
\label{tab:text_generation_alternariol}
\resizebox{\columnwidth}{!}{%
\begin{tabular}{@{}llllllllllll@{}}
\toprule
Prompt                                              & GPT-2 & GPT-3                                                                                                       & GPT-neo                              & Bloom                      & BioGPT                                 & BioGPT-Large                           & ChatGPT                                                                                                            & GPT-4                                                                                                              & Llama 2-7B                                                                                         & Llama 2-13B                                                                                                                                                                                                                                                                   & Llama 2-70B                                                                                                \\ \midrule
Alternariol is produced by                          &       & \textit{\textbf{Alternaria alternata}}                                                                      & \textit{}                            & \textit{}                  & \textit{\textbf{Alternaria alternata}} & \textit{\textbf{Phomopsis amygdali}}   & \textit{\textbf{Alternaria}}                                                                                       & \textit{\textbf{\begin{tabular}[c]{@{}l@{}}Alternaria alternata, \\ A. tenuissima, A. mali\end{tabular}}}          & \textit{\begin{tabular}[c]{@{}l@{}}Amanita phalloides, \\ A. phalloides var. virosa\end{tabular}}  & \textit{\textbf{\begin{tabular}[c]{@{}l@{}}Alternaria alternata, \\ A. arborescens, \\ A. brassicicola, \\ A. carotae, A. dauci, \\ A. longipes, A. radicina, \\ A. tenuissima, \\ A. triticina, A. zinniae\end{tabular}}}                                                    & \textit{\textbf{Alternaria alternata}}                                                                     \\
Alternariol is produced by the fungus               &       & \textit{\textbf{Alternaria alternata}}                                                                      & \textit{}                            & \textit{Penicillium acnes} & \textit{\textbf{Alternaria alternata}} & \textit{\textbf{Phomopsis amygdali}}   & \textit{\textbf{Alternaria}}                                                                                       & \textit{\textbf{\begin{tabular}[c]{@{}l@{}}Alternaria alternata, \\ A. tenuissima\end{tabular}}}                   & \textit{}                                                                                          & \textit{\textbf{Alternaria alternata}}                                                                                                                                                                                                                                        & \textit{\textbf{Alternaria alternata}}                                                                     \\
Alternariol is produced by fungi                    &       & \textit{\textbf{\begin{tabular}[c]{@{}l@{}}Alternaria, \\ Fusarium, \\ Penicillium\end{tabular}}}           & \textit{\begin{tabular}[c]{@{}l@{}}Ascomycota: \\ Tremellomycetes\end{tabular}}  & \textit{}                  & \textit{\textbf{}}                     & \textit{\textbf{}}                     & \textit{\textbf{Alternaria}}                                                                                       & \textit{\textbf{\begin{tabular}[c]{@{}l@{}}Alternaria alternata, \\ A. tenuissima, \\ A. infectoria\end{tabular}}} & \textit{\textbf{Alternaria}}                                                                       & \textit{\textbf{Alternaria}}                                                                                                                                                                                                                                                  & \textit{\textbf{Alternaria}}                                                                               \\
Alternariol is produced by fungi, such as           &       & \textit{\textbf{Alternaria}}                                                                                & \textit{\textbf{Aspergillus}}        & \textit{}                  & \textit{\textbf{Alternaria alternata}} & \textit{\textbf{Alternaria alternata}} & \textit{\textbf{\begin{tabular}[c]{@{}l@{}}Alternaria alternata, \\ A. tenuissima, \\ A. infectoria\end{tabular}}} & \textit{\textbf{\begin{tabular}[c]{@{}l@{}}Alternaria alternata, \\ A. tenuissima, \\ A. infectoria\end{tabular}}} & \textit{\begin{tabular}[c]{@{}l@{}}Penicillium verrucosum, \\ Aspergillus versicolor\end{tabular}} & \textit{\textbf{\begin{tabular}[c]{@{}l@{}}Alternaria alternata, \\ A. tenuissima\end{tabular}}}                                                                                                                                                                              & \textit{\textbf{Alternaria alternata}}                                                                     \\
Alternariol is isolated from                        &       & \textit{\textbf{Alternaria}}                                                                                & \textit{}                            & \textit{}                  & \textit{\textbf{}}                     & \textit{\textbf{Phomopsis amygdali}}   & \textit{\textbf{Alternaria}}                                                                                       & \textit{\textbf{Alternaria}}                                                                                       & \textit{\textbf{Alternaria}}                                                                       & \textit{\textbf{\begin{tabular}[c]{@{}l@{}}Alternaria alternata, \\ A. tenuissima\end{tabular}}}                                                                                                                                                                              & \textit{\textbf{\begin{tabular}[c]{@{}l@{}}Alternaria alternata, \\ Aureobasidium \\pullulans\end{tabular}}} \\
Alternariol originally isolated from                &       & \textit{\textbf{Alternaria alternata}}                                                                      & \textit{Aspergillus terreus}         & \textit{}                  & \textit{\textbf{}}                     & \textit{Phellinus linteus}             & \textit{\textbf{Alternaria tenuissima}}                                                                            & \textit{\textbf{Alternaria alternata}}                                                                             & \textit{\textbf{Alternaria alternata}}                                                             & \textit{\textbf{Alternaria alternata}}                                                                                                                                                                                                                                        & \textit{\textbf{Alternaria}}                                                                               \\
Alternariol is isolated from the fungus             &       & \textit{\textbf{Alternaria alternata}}                                                                      & \textit{}                            & \textit{}                  & \textit{\textbf{Alternaria alternata}} & \textit{\textbf{Phomopsis amygdali}}   & \textit{\textbf{Alternaria}}                                                                                       & \textit{\textbf{\begin{tabular}[c]{@{}l@{}}Alternaria alternata, \\ A. tenuissima\end{tabular}}}                   & \textit{\textbf{}}                                                                                 & \textit{\textbf{Alternaria alternata}}                                                                                                                                                                                                                                        & \textit{\textbf{Alternaria alternata}}                                                                     \\
Alternariol is isolated from fungi                  &       & \textit{\textbf{}}                                                                                          & \textit{}                            & \textit{}                  & \textit{\textbf{Alternaria}}           & \textit{\textbf{}}                    & \textit{\textbf{Alternaria}}                                                                                       & \textit{\textbf{\begin{tabular}[c]{@{}l@{}}Alternaria alternata, \\ A. tenuissima\end{tabular}}}                   & \textit{\begin{tabular}[c]{@{}l@{}}Penicillium verrucosum, \\ Aspergillus niger\end{tabular}}      & \textit{\textbf{\begin{tabular}[c]{@{}l@{}}Alternaria alternata, \\ A. arborescens, \\ A. brassicicola, \\ A. carotae, A. dauci, \\ A. eres, A. fennica, \\ A. iowensis, A. longipes, \\ A. mali, A. mirabilis, \\ A. radicina, \\ A. tenuissima, \\A. triticina\end{tabular}}} & \textit{\textbf{Alternaria}}                                                                               \\
Alternariol is isolated from fungi, such as         &       & \textit{\textbf{\begin{tabular}[c]{@{}l@{}}Alternaria alternata, \\ A. tenuissima, A. solani\end{tabular}}} & \textit{}                            & \textit{}                  & \textit{\textbf{Alternaria alternata}} & \textit{\textbf{Phomopsis}}            & \textit{\textbf{\begin{tabular}[c]{@{}l@{}}Alternaria alternata, \\ A. tenuissima, \\ A. infectoria\end{tabular}}} & \textit{\textbf{\begin{tabular}[c]{@{}l@{}}Alternaria alternata, \\ A. solani, A. tenuissima\end{tabular}}}        & \textit{\begin{tabular}[c]{@{}l@{}}Penicillium verrucosum, \\ Aspergillus niger\end{tabular}}      & \textit{\textbf{Alternaria}}                                                                                                                                                                                                                                                  & \textit{\textbf{}}                                                                                         \\
Alternariol is a natural product found in           &       & \textit{\textbf{}}                                                                                          &                                      &                            & \textbf{}                              & \textbf{}                              & \textit{\textbf{Alternaria}}                                                                                       & \textit{\textbf{Alternaria}}                                                                                       &                                                                                                    & \textit{\textbf{}}                                                                                                                                                                                                                                                            & \textit{\textbf{Alternaria alternata}}                                                                     \\
\begin{tabular}[c]{@{}l@{}}Alternariol is a natural chemical\\ compound found in \end{tabular}&       & \textbf{}                                                                                                   &                                      &                            & \textbf{}                              & \textbf{}                              & \textit{\textbf{Alternaria}}                                                                                       & \textit{\textbf{Alternaria}}                                                                                       &                                                                                                    & \textit{\textbf{Alternaria alternata}}                                                                                                                                                                                                                                        & \textit{\textbf{Alternaria}}                                                                               \\ \bottomrule
\end{tabular}%
}
\end{table}
\end{landscape}
\pagebreak

\begin{landscape}
\begin{table}[]
\caption{Table \thetable. Generated fungus name in Task 2: \textit{chemical-fungus relation determination via entity generation task} for one chemical compound \textit{Verrucarin A} for 11 prompts (P5-P15). Fungus names that are generated correctly in bold.}
\refstepcounter{app_table} 
\label{tab:text_generation_verrucarina}
\resizebox{\columnwidth}{!}{%
\begin{tabular}{@{}llllllllllll@{}}
\toprule
Prompt                                               & GPT-2     & GPT-3                          & GPT-neo                                                                         & Bloom                         & BioGPT                           & BioGPT-Large                    & ChatGPT                                                                                                    & GPT-4                                    & Llama 2-7B                                                                                    & Llama 2-13B                     & Llama 2-70B                    \\ \midrule
Verrucarin A is produced by                          & \textit{} & \textit{Verrucaria nigrescens} & \textit{}                                                                       & \textit{}                     & \textit{Aspergillus niger}       & \textit{Penicillium verrucosum}                     & \textit{\textbf{Myrothecium verrucaria}}                                                                   & \textit{\textbf{Myrothecium verrucaria}} & \textit{\begin{tabular}[c]{@{}l@{}}Penicillium verrucosum, \\ Aspergillus niger\end{tabular}} & \textit{}                       & \textit{Verrucaria}            \\
Verrucarin A is produced by the fungus               & \textit{} & \textit{Verrucaria nigrescens} & \textit{}                                                                       & \textit{Penicillium expansum} & \textit{Aspergillus niger}       & \textit{Penicillium verrucosum}                     & \textit{\textbf{Myrothecium verrucaria}}                                                                   & \textit{\textbf{Myrothecium verrucaria}} & \textit{}                                                                                     & \textit{}                       & \textit{}                      \\
Verrucarin A is produced by fungi                    & \textit{} & \textit{Verrucaria}            & \textit{Neurospora crassa}                                                      & \textit{}                     & \textit{}                        & \textit{Penicillium}            & \textit{\textbf{Myrothecium verrucaria}}                                                                   & \textit{\textbf{Myrothecium verrucaria}} & \textit{Verrucaria}                                                                           & \textit{Verrucaria}             & \textit{Verrucaria}            \\
Verrucarin A is produced by fungi, such as           & \textit{} & \textit{Verrucaria nigrescens} & \begin{tabular}[c]{@{}l@{}}\textit{Penicillium roqueforti}, \\ \textit{Penicillium verrucarinum}\end{tabular}                       & \textit{}                     & \textit{Aspergillus terreus}     & \textit{Penicillium verrucosum} & \textbf{\begin{tabular}[c]{@{}l@{}}\textit{Myrothecium verrucaria}, \\ \textit{Stachybotrys chartarum}, \\ \textit{Fusarium oxysporum}, \\ \textit{Alternaria alternata}\end{tabular}} & \textit{\textbf{Myrothecium verrucaria}} & \textit{Penicillium verrucosum}                                                               & \textit{Aspergillus}            & \textit{}                      \\
Verrucarin A is isolated from                        & \textit{} & \textit{Verrucaria maura}      & \textit{}                                                                       & \textit{}                     & \textit{}                        & \textit{Penicillium verrucosum} & \textit{\textbf{Myrothecium verrucaria}}                                                                   & \textit{\textbf{Myrothecium verrucaria}} & \textit{Verrucaria}                                                                           & \textit{Verrucaria}             & \textit{Aspergillus}           \\
Verrucarin A originally isolated from                & \textit{} & \textit{Verrucaria nigrescens} & \textit{}                                                                       & \textit{}                     & \textit{}                        & \textit{Penicillium verrucosum}                    & \textit{\textbf{Myrothecium verrucaria}}                                                                   & \textit{\textbf{Myrothecium verrucaria}} & \textit{}                                                                                     & \textit{Penicillium verrucarum} & \textit{}                      \\
Verrucarin A is isolated from the fungus             & \textit{} & \textit{Verrucaria nigrescens} & \textit{}                                                                       & \textit{Penicillium expansum} & \textit{Aspergillus  versicolor} & \textit{Penicillium verrucosum} & \textit{\textbf{Myrothecium verrucaria}}                                                                   & \textit{\textbf{Myrothecium verrucaria}} & \textit{}                                                                                     & \textit{}                       & \textit{}                      \\
Verrucarin A is isolated from fungi                  & \textit{} & \textit{Verrucaria}            & \textit{}                                                                       & \textit{}                     & \textit{Aspergillus}             & \textit{}                       & \textit{\textbf{Myrothecium verrucaria}}                                                                   & \textit{\textbf{Myrothecium verrucaria}} & \textit{Verrucaria}                                                                           & \textit{Verrucaria}             & \textit{Verrucaria}            \\
Verrucarin A is isolated from fungi, such as         & \textit{} & \textit{Verrucaria maura}      & \begin{tabular}[c]{@{}l@{}}\textit{Conchocalyx irregularis}, \\ \textit{Aspergillus ochraceus}, \\ \textit{Stachybotrys chartarum}\end{tabular} & \textit{}                     & \textit{Aspergillus terreus}     & \textit{Penicillium verrucosum} & \textbf{\begin{tabular}[c]{@{}l@{}}\textit{Myrothecium verrucaria}, \\ \textit{Stachybotrys chartarum}, \\ \textit{Fusarium oxysporum}, \\ \textit{Alternaria alternata}\end{tabular}} & \textit{\textbf{Myrothecium verrucaria}} & \textit{}                                                                                     & \textit{}                       & \textit{Verrucaria nigrescens} \\
Verrucarin A is a natural product found in           & \textit{} & \textit{Verrucaria}            & \textit{}                                                                       & \textit{}                     & \textit{}                        & \textit{Penicillium verrucosum} & \textbf{\begin{tabular}[c]{@{}l@{}}\textit{Myrothecium verrucaria}, \\ \textit{Stachybotrys chartarum}, \\ \textit{Fusarium oxysporum}, \\ \textit{Alternaria alternata}\end{tabular}} & \textit{\textbf{Myrothecium}}            & \textit{}                                                                                     & \textit{Verrucaria}             & \textit{}                      \\
Verrucarin A is a natural chemical compound found in & \textit{} & \textit{Verrucaria}            & \textit{}                                                                       & \textit{}                     & \textit{}                        & \textit{Penicillium verrucosum} & \textbf{\begin{tabular}[c]{@{}l@{}}\textit{Myrothecium verrucaria}, \\ \textit{Stachybotrys chartarum}, \\ \textit{Fusarium oxysporum}, \\ \textit{Alternaria alternata}\end{tabular}} & \textit{\textbf{Myrothecium}}            & \textit{Verrucaria}                                                                           & \textit{Verrucaria}             & \textit{}                      \\ \bottomrule
\end{tabular}%
}
\end{table}
\end{landscape}

\pagebreak 

\begin{table}[]
\centering
\caption{Table \thetable. Performance in the Task 2: \textit{chemical-fungus relation determination via entity generation} task. Evaluation according to the framework (Fig. \ref{fig:framework}). 10 chemicals, 11 prompts (P5-P15), 110 outputs per model in total. Analysis of the occurrence of the most common fungus in chemical compound-fungus relation \textit{Aspergillus} in tested prompts.}
\refstepcounter{app_table} 
\label{tab:text_generation_relation}
\resizebox{\columnwidth}{!}{%
\begin{tabular}{@{}lllll@{}}
\toprule
Model        & \begin{tabular}[c]{@{}l@{}}Occurrence of any fungus name \\ in the answer / \\ tested prompts (\%)\end{tabular} & \begin{tabular}[c]{@{}l@{}}Occurrence of \textit{Aspergillus} \\ in the aswer  / \\ occurrence of any fungus name \\ in the the answer (\%)\end{tabular} & \begin{tabular}[c]{@{}l@{}}Factual occurrence of \\ \textit{Aspergillus} in \\ the answer / occurrence \\ of fungus name in \\ the answer (\%)\end{tabular} & \begin{tabular}[c]{@{}l@{}}Factual occurrence of \\ \textit{Aspergillus} in \\ the answer / occurrence \\ of \textit{Aspergillus} \\ in the answer (\%)\end{tabular} \\ \midrule
GPT-2        & 0/110 (0)                                                                                                      & 0/0 (0)                                                                                                                                                 & 0/0 (0)                                                                                                                                                    & 0/0 (0)                                                                                                                                                         \\
GPT-3        & 56/110 (51)                                                                                                    & 21/56 (37.5)                                                                                                                                            & 14/56 (25)                                                                                                                                                 & 14/21 (66.7)                                                                                                                                                    \\
GPT-neo      & 28/110 (25.5)                                                                                                  & 11/28 (39.3)                                                                                                                                            & 4/28 (14.3)                                                                                                                                                & 4/11 (36.4)                                                                                                                                                     \\
Bloom        & 12/110 (11)                                                                                                    & 3/12 (25)                                                                                                                                               & 2/12 (16.7)                                                                                                                                                & 2/3 (66.7)                                                                                                                                                      \\
BioGPT       & 57/110 (51.8)                                                                                                  & 45/57 (78.9)                                                                                                                                            & 13/57 (22.8)                                                                                                                                               & 13/45 (28.9)                                                                                                                                                    \\
BioGPT-large & 74/110 (67.3)                                                                                                  & 20/74 (27)                                                                                                                                              & 13/74 (17.6)                                                                                                                                               & 13/20 (65)                                                                                                                                                      \\
ChatGPT      & 101/110 (91.8)                                                                                                 & 21/101 (20.8)                                                                                                                                           & 12/101 (11.9)                                                                                                                                              & 12/21 (57.1)                                                                                                                                                    \\
GPT-4        & 77/110 (70)                                                                                                    & 21/77 (27.3)                                                                                                                                            & 15/77 (19.5)                                                                                                                                               & 15/21 (71.4)                                                                                                                                                    \\
Llama 2-7B   & 36/110 (32.7)                                                                                                  & 6/36 (16.7)                                                                                                                                             & 0/36 (0)                                                                                                                                                   & 0/6 (0)                                                                                                                                                         \\
Llama 2-13B  & 49/110 (44.5)                                                                                                  & 11/49 (22.5)                                                                                                                                            & 1/49 (2)                                                                                                                                                   & 1/11 (9.1)                                                                                                                                                      \\
Llama 2-70B  & 62/110 (56.4)                                                                                                  & 17/62 (27.4)                                                                                                                                            & 7/62 (11.3)                                                                                                                                                & 7/17 (41.2)                                                                                                                                                     \\ \bottomrule
\end{tabular}%
}
\end{table}

\begin{landscape}
\begin{table}[]
\caption{Table \thetable. Generated fungus name in Task 2: \textit{chemical-fungus relation determination via entity generation task} for one chemical compound \textit{Ergosterol} for 11 prompts (P5-P15). Fungus names that are generated correctly in bold.}
\refstepcounter{app_table} 
\label{tab:text_generation_ergosterol}
\resizebox{\columnwidth}{!}{%
\begin{tabular}{@{}lllllllllllll@{}}
\toprule
Prompt                                             & GPT-2     & GPT-3                             & GPT-neo                       & Bloom                         & BioGPT                       & BioGPT-Large                                                             & ChatGPT                                                                                             & GPT-4     & Llama 2-7B & Llama 2-13B & Llama 2-70B                 \\ \midrule
Ergosterol is produced by                          & \textit{} & \textit{}                         & \textit{}                     & \textit{}                     & \textit{\begin{tabular}[c]{@{}l@{}}Aspergillus \\ terreus \end{tabular}} & \textit{\begin{tabular}[c]{@{}l@{}}Ganoderma \\lucidum \end{tabular}}                                               & \textit{}                                                                                           & \textit{} & \textit{}  & \textit{}   & \textit{}                   \\
Ergosterol is produced by the fungus               & \textit{} & \begin{tabular}[c]{@{}l@{}}\textit{Saccharomyces} \\ \textit{cerevisiae}\end{tabular} & \begin{tabular}[c]{@{}l@{}}\textit{Aspergillus} \\ \textit{nidulans}\end{tabular} & \begin{tabular}[c]{@{}l@{}}\textit{Aspergillus} \\ \textit{niger}\end{tabular}    & \textit{\begin{tabular}[c]{@{}l@{}}Aspergillus \\ terreus \end{tabular}} & \textit{\begin{tabular}[c]{@{}l@{}}Ganoderma \\ lucidum \end{tabular}}                                              & \textit{}                                                                                           & \textit{} & \textit{}  & \textit{}   & \textit{}                   \\
Ergosterol is produced by fungi                    & \textit{} & \textit{}                         & \textit{}                     & \textit{}                     & \textit{}                    & \textit{}                                                                & \textit{}                                                                                           & \textit{} & \textit{}  & \textit{}   & \textit{}                   \\
Ergosterol is produced by fungi, such as           & \textit{} & \textit{}                         & \textit{Emericella}           & \textit{}                     & \textit{\begin{tabular}[c]{@{}l@{}}Aspergillus \\ terreus \end{tabular}}  & \textit{\begin{tabular}[c]{@{}l@{}}Ganoderma \\ lucidum \end{tabular}}                                               & \begin{tabular}[c]{@{}l@{}}\textit{Aspergillus}, \\ \textit{Penicillium}, \\ \textit{Fusarium}\end{tabular}                                                         & \textit{} & \textit{}  & \textit{}   & \textit{}                   \\
Ergosterol is isolated from                        & \textit{} & \textit{}                         & \textit{}                     & \textit{}                     & \textit{}                    & \textit{\begin{tabular}[c]{@{}l@{}}Ganoderma \\ lucidum \end{tabular}}                                               & \textit{}                                                                                           & \textit{} & \textit{}  & \textit{}   & \textit{}                   \\
Ergosterol originally isolated from                & \textit{} & \textit{}                         & \textit{}                     & \textit{}                     & \textit{}                    & \textit{\begin{tabular}[c]{@{}l@{}}Ganoderma \\ lucidum \end{tabular}}                                               & \textit{\begin{tabular}[c]{@{}l@{}}Claviceps \\purpurea\end{tabular}}                                                                         & \textit{} & \textit{}  & \textit{}   & \textit{}                   \\
Ergosterol is isolated from the fungus             & \textit{} & \textit{}                         & \textit{}                     & \begin{tabular}[c]{@{}l@{}}\textit{Penicillium} \\ \textit{expansum}\end{tabular} & \textit{\begin{tabular}[c]{@{}l@{}}Aspergillus \\ terreus\end{tabular}} & \textit{\begin{tabular}[c]{@{}l@{}}Ganoderma \\ lucidum \end{tabular}}                                               & \textit{\begin{tabular}[c]{@{}l@{}}Claviceps \\purpurea\end{tabular}}                                                                         & \textit{} & \textit{}  & \textit{}   & \textit{}                   \\
Ergosterol is isolated from fungi                  & \textit{} & \textit{}                         & \textit{}                     & \textit{}                     & \textit{Aspergillus}         & \textit{}                                                                & \textit{}                                                                                           & \textit{} & \textit{}  & \textit{}   & \textit{}                   \\
Ergosterol is isolated from fungi, such as         & \textit{} & \textit{}                         & \textit{}                     & \textit{}                     & \textit{\begin{tabular}[c]{@{}l@{}}Aspergillus \\ terreus \end{tabular}} & \begin{tabular}[c]{@{}l@{}}\textit{\begin{tabular}[c]{@{}l@{}}Ganoderma \\ lucidum \end{tabular}}, \\ \textit{\begin{tabular}[c]{@{}l@{}}Ganoderma \\ applanatum \end{tabular}}, \\ \textit{\begin{tabular}[c]{@{}l@{}}Ganoderma \\ neojaponicum \end{tabular}}\end{tabular} & \begin{tabular}[c]{@{}l@{}}\textit{\begin{tabular}[c]{@{}l@{}}Candida \\ albicans\end{tabular}}, \\ \textit{\begin{tabular}[c]{@{}l@{}}Aspergillus \\ fumigatus \end{tabular}}, \\ \textit{\begin{tabular}[c]{@{}l@{}}Saccharomyces \\cerevisiae \end{tabular}}, \\ \textit{\begin{tabular}[c]{@{}l@{}}Penicillium \\ chrysogenum\end{tabular}}\end{tabular} & \textit{} & \textit{}  & \textit{}   & \textit{\begin{tabular}[c]{@{}l@{}}Cordyceps \\ sinensis \end{tabular}} \\
Ergosterol is a natural product found in           & \textit{} & \textit{}                         & \textit{}                     & \textit{}                     & \textit{}                    & \textit{}                                                                & \textit{}                                                                                           & \textit{} & \textit{}  & \textit{}   & \textit{}                   \\
\begin{tabular}[c]{@{}l@{}}Ergosterol is a natural chemical compound \\found in \end{tabular}& \textit{} & \textit{}                         & \textit{}                     & \textit{}                     & \textit{}                    & \textit{}                                                                & \textit{}                                                                                           & \textit{} & \textit{}  & \textit{}   & \textit{}      \\ \bottomrule          
\end{tabular}%
}
\end{table}
\end{landscape}








\end{document}